\documentclass[letterpaper, 10 pt, conference]{ieeeconf}  
\usepackage{amsmath,amssymb,amsfonts}
\usepackage{graphicx}
\usepackage{caption}
\usepackage{subcaption}
\usepackage{textcomp}
\usepackage[dvipsnames]{xcolor}
\usepackage{booktabs}
\usepackage[final]{changes}

\usepackage{makecell}
\usepackage{mathrsfs}

\usepackage{multirow}

\usepackage{style/ynl_handout}

\IEEEoverridecommandlockouts                              

\title{\LARGE \bf
PC7 Model: Dynamics of Expanding and Contracting Soft Robots\\using Piece-wise Constant 7D Deformation Based on Similarity Transformation Group
}

\author{
Yamato Kudo, Taiki Ishigaki, Ko Ayusawa and Ko Yamamoto
\thanks{}
\thanks{
}%
\thanks{
}%
}

\begin{document}

\maketitle
\thispagestyle{empty}
\pagestyle{empty}

\begin{abstract}
This paper proposes a novel soft robot model named piece-wise constant 7D deformation (PC7) model that can represent expanding and contracting deformation of, for example, artificial muscle actuator.
In the proposed model, we newly define the 7D deformation vector based on SIM(3) -- 3D similarity transformation group -- whereas a 6D deformation vector was defined for Cosserat rod based on SE(3) -- the special Euclid group -- in the piece-wise constant strain (PCS) model proposed in the previous study.
The additional dimension of the 7D deformation vector represents expanding and contracting deformation, which comes from the scaling parameter in SIM(3).
Common mathematical properties between SE(3) and SIM(3) allow us to directly extend the kinematics and dynamics of the PCS model to those of the PC7 model.
In particular, we derive the dynamical equation of the PC7 model based on Euler-Poincar\'e equation in a mathematically precise manner.
We validate the proposed model through dynamical simulations assuming a thin cylinder structure or McKibben-type artificial muscle actuator.
\end{abstract}

\section{Introduction}
Artificial muscle actuator including McKibben-type one \cite{tondu2012modelling} is often employed in bio-inspired soft robots \cite{laschi2012soft,marcheseandrew2014autonomous} and musculo-skeletal robots \cite{niiyama2012biomechanical,kurumaya2016musculoskeletal}.
High flexibility and backdrivability of such an actuator contribute to not only the adaptability to unknown environment but also dynamic motion ability.
A lot of studies have investigated kinematic and dynamical modeling of soft and continuous deformation.
Finite element method (FEM) is a representative one that allows detailed modeling of a complicated 3D shape, as seen in artificial muscle actuator.
In general, however, computational cost of FEM becomes higher as the number of meshes increases.
It usually requires some assumptions to make the computational speed fast \cite{navarro2023open}.

\begin{figure}[tbp]
  \begin{center}
    \includegraphics[width=0.9\hsize]{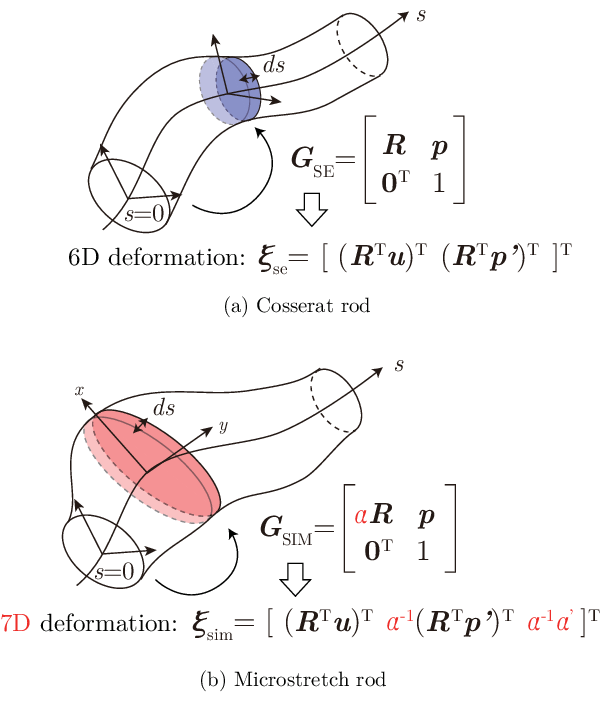}   
  \end{center}
  \caption{Schematic diagram of (a) Cosserat rod and (b) microstretch rod \cite{eringen2012microcontinuum}. In the Cosserat rod, the configuration of miscrostructre is represented by SE(3), and a 6D deformation $\bm{\xi}_{\mathfrak{se}}$ is defined by $\mathfrak{se}(3)$ \cite{renda2018discrete}. In the microstretch rod, the configuration can be represented by SIM(3), and we can newly define a 7D deformation $\bm{\xi}_{\mathfrak{sim}}$ by $\mathfrak{sim}(3)$, as proposed in this study.}
  \label{fig:cosserat_microstretch_rods}
\end{figure}

There is another approach focusing on soft rod structure.
In particular, discretization and piece-wise constant assumption are often employed \cite{Webster2010,Katzschmann2019,renda2018discrete}, which results in a formulation similar to that proposed in traditional rigid robotics.
Renda et al. \cite{renda2018discrete} proposed a discrete Cosserat rod model, also referred to as Piece-wise Constant Strain (PCS) model.
Cosserat continuum theory assumes that a miscrosructure in a continuum is represented as a rigid body whereas classical continuum theory assumes that an elemet  is infinitesimal particle.
Therefore, its configuration can be represented by special Euclid group, SE(3), as shown in \figref{fig:cosserat_microstretch_rods}(a).
In the PCS model, a 6D deformation vector was defined based on $\mathfrak{se}(3)$, which is Lie algebra of SE(3), and the dynamical equation was derived based on Euler-Poincar\'e equation (or simply Poincar\'e equation) \cite{Marle2013,boyer2017poincare}, especially for a finite number of segments that are obtained by discretizing a continuum.
The obtained formulation is compatible with the traditional rigid-link system because the traditional kinematic chain is also described by SE(3) and $\mathfrak{se}(3)$ \cite{featherstone2008rigid}.
This compatibility enabled an extension to hybrid-link system integrating rigid-link and PCS models \cite{ishigaki2021dynamics,mathew2022sorosim}.
Moreover, framework for calculating analytical derivatives of the PCS and hybrid-link models have been recently investigated \cite{Ishigaki2024,ishigaki2025unified,Mathew2025}, based on mathematical properties of Lie group.

Although the PCS model or Cosserat rod theory cannot directly deal with expanding and contracting deformation seen in artificial muscle actuator, several extensions have been proposed \cite{angles2019viper,Tobin2025}.
Angels et al. \cite{angles2019viper} pointed that such expanding and contracting deformation can be described by similarity transformation group, SIM(3), which includes uniform scaling of the coordinate system.
However, their approach relies on a Position-Based Dynamics (PBD) framework tailored for visually plausible computer graphics, rather than deriving mathematically strict equations of motion based on Euler-Poincaré equations.
Recently, Tobin et al. \cite{Tobin2025} extended the Cosserat rod model by introducing additional degrees of freedom \added{(DOFs)} to parameterize in-plane stretch and shear of the cross-section. However, their framework \replaced{was}{is} based on the minimization of total potential energy and is thus limited to static or quasi-static analysis. \deleted{(この点は後ほどSection IIIで詳しく述べるので、ここでは省略) Furthermore, because their parameterization deviates from the strict Lie group formulation, extending their model to macroscopic dynamics would lead to excessive mathematical complexity.}

To address these limitations, we newly define the 7D deformation vector based on SIM(3), as shown in \figref{fig:cosserat_microstretch_rods}(b). The additional dimension of the 7D deformation vector represents expanding and contracting deformation perpendicular to the rod central axis, which comes from the scaling parameter of SIM(3).
By preserving the Lie group structure, we propose a novel soft robot model named {\it piece-wise constant 7D deformation (PC7)} model. This model directly extends the kinematics and dynamics of the PCS model based on common mathematical properties between SE(3) and SIM(3), allowing us to derive the macroscopic dynamical equation of the PC7 model based on Poincar'e equation without mathematical inconsistency.
To showcase the applicability of this theoretical framework, we present dynamical simulations of expanding and contracting structures. These examples demonstrate how the PC7 model computes macroscopic dynamic behaviors based on the derived equations. 

The rest of this paper is organized as follows. 
Section \ref{sect:cosserat_rod} reviews the derivation of Poincar\'e equations for Cosserat rod.
Then, Section \ref{sect:microstretch_rod} formulates Poincar\'e equation of miscrostretch rod \cite{eringen2012microcontinuum}, which is an extension of Cosserat rod considering expanding and contracting deformation perpendicular to the rod central axis.
Based on this formulation, we propose the PC7 model and formulate its dynamical equation in Section \ref{sect:pc7_model}.
\added{The PC7 model dynamics can be easily resolved with a holonomic constraint in a manner similar to the traditional rigid-body dynamics.
Section \ref{sect:ex_constrained_mechanisms} presents two examples of such constraints seen in a thin-walled expanding cylinder and McKibben-type artificial muscle actuator.}
Section \ref{sect:simulation} provides quantitative validations through dynamical simulations of a thin-walled cylinder and McKibben-type artificial muscle actuator, comparing the simulation results of the PC7 model with those of FEM, and also with an experimental data reported in a previous study.

Hereafter, $t$ denotes time, and $s$ denotes a spatial variable such as the center-axis coordinate of a flexible rod shown in \figref{fig:cosserat_microstretch_rods}.
When it is clear that a variable is a function of $t$ or $s$, we omit either of them.
Also, we use the following notations:
\begin{itemize}
    \item $\dot{*} := \pdif{*}{t}$ and $*^{\prime} := \pdif{*}{s}$ are time- and spatial-derivatives of a quantity $\ast$, respectively.
    \item $\bm{0} \in\mathbb{R}^3$ and $\bm{O} \in\mathbb{R}^{3\times3}$ are zero vector and matrix, respectively;
    \item $\bm{E} \in\mathbb{R}^{3\times3}$ is the unit matrix;
    \item $\mathrm{diag}\{ * \}$ is a (block) diagonal matrix consisting of elements $\ast$.
    \item $\bm{e}_6 :=\begin{bmatrix}
        \bm{0}^T & 0 & 0 & 1 & 0 
    \end{bmatrix}^T$ is a 7D unit vector in which the sixth element is 1.
    \item $\bm{e}_7:=\begin{bmatrix}
        \bm{0}^T & 0 & 0 & 0 & 1 
    \end{bmatrix}^T$ is a 7D unit vector in which the seventh element is 1.
\end{itemize}

The basics of 3D Lie group calculations are summarized as follows.
SO(3) is the 3D special orthogonal group, and $\mathfrak{so}(3)$ is its Lie algebra.
        Given $\bm{a} = [ a_x \ a_y \ a_z ]^T$, $[\bm{a}\times_3] \in \mathfrak{so}(3)$ is defined as
\begin{align}
    \label{eq:cross_product_mat}
    [\bm{a} \times_3] := 
    \begin{bmatrix}
    0 & -a_z & a_y \\
    a_z & 0 & -a_x\\
    -a_y & a_x & 0
    \end{bmatrix} .
\end{align}

SE(3) is the 3D special Euclid group, and $\mathfrak{se}(3)$ is its Lie algebra.
    Given $\bm{R} \in$ SO(3) and $\bm{p} \in \mathbb{R}^3$, $\bm{G}_{\rm{SE}}(\bm{R}, \bm{p}) \in$ SE(3) is represented as
    \begin{align}
        \bm{G}_{\rm{SE}}(\bm{R}, \bm{p})
        =
        \begin{bmatrix}
            \bm{R} & \bm{p} \\
            \bm{0}^T & 1
        \end{bmatrix} .
    \end{align}
    Its adjoint representation is given as
    \begin{align}
    \bm{A}_{\rm{SE}}
    :=
    \begin{bmatrix}
        \bm{R} & \bm{O} \\
        [\bm{p} \times]\bm{R} & \bm{R}
    \end{bmatrix} .
    \end{align}
    Given $\bm{a}$ and $\bm{l} \in \mathbb{R}^3$, $[\bm{x}_{\mathfrak{se}}\times_4]$ and $[\bm{x}_{\mathfrak{se}}\times_6]$ are defined for a 6D vector $\bm{x}_{\mathfrak{se}} = [\bm{a}^T \ \bm{l}^T]^T$ as 
\begin{align}
 [ \bm{x}_{\mathfrak{se}} \times_4 ] &:=
 \begin{bmatrix}
  [ \bm{a} \times_3 ] & \bm{l} \\
  \bm{0}^T & 0
 \end{bmatrix} \in \mathfrak{se}(3),
 \\
    [ \bm{x}_{\mathfrak{se}} \times_6 ] &:=
    \begin{bmatrix}
        [\bm{a}\times_3] & \bm{O} \\
        [\bm{l}\times_3] & [\bm{a}\times_3]
    \end{bmatrix}.
\end{align}
    Note that $[\bm{x}\times_6]$ is the adjoint representation of $\mathfrak{se}(3)$.

SIM(3) is the 3D similarity transformation group, and $\mathfrak{sim}(3)$ is its Lie algebra.
        Given $\bm{R} \in$ SO(3), $\bm{p} \in \mathbb{R}^3$ and $\alpha > 0$, $\bm{G}_{\rm SIM}(\bm{R}, \bm{p}, \alpha) \in$ SIM(3) and its adojoint representation are represented as
\begin{subequations}
\begin{align}
    \label{eq:similarity_trans_mat}
    \bm{G}_{\rm SIM}(\bm{R}, \bm{p}, \alpha)
    =
    \begin{bmatrix}
        \alpha \bm{R} & \bm{p} \\
        \bm{0}^T & 1
    \end{bmatrix}, \mbox{and} 
    \\
    \bm{A}_{\rm SIM}
    =
    \begin{bmatrix}
    \bm{R} & \bm{O} & \bm{0} \\
    [\bm{p} \times] \bm{R} & \alpha \bm{R} & -\bm{p} \\
    \bm{0}^T & \bm{0}^T & 1
    \end{bmatrix} .
\end{align}
\end{subequations}
        Given $\bm{a}, \bm{l} \in \mathbb{R}^3$ and $\lambda \in \mathbb{R}$, $[\bm{x}_{\mathfrak{sim}}\times_4]$ and $[\bm{x}_{\mathfrak{sim}}\times_7]$ are defined for a 7D vector $\bm{x}_{\mathfrak{sim}} = [\bm{a}^T \ \bm{l}^T \lambda]^T$  as 
\begin{subequations}
\begin{align}
    \label{eq:sim3_algbr}
    [\bm{x}_{\mathfrak{sim}}\times_4] &= 
    \begin{bmatrix}
        \lambda\bm{E}+[\bm{a}\times_3] & \bm{l}\\
        \bm{0}^T & 0
    \end{bmatrix} \in \mathfrak{sim}(3),
    \\
    [\bm{x}_{\mathfrak{sim}}\times_7] &= 
    \begin{bmatrix}
        [\bm{a}\times_3] & \bm{O} & \bm{0}\\
        [\bm{l}\times_3] & \lambda\bm{E}+[\bm{a}\times_3] & -\bm{l}\\
        \bm{0}^T & \bm{0}^T & 0
    \end{bmatrix}.
\end{align}
\end{subequations}

We also define a scaling matrix \added{and its adjoint representation} as follows:
\begin{subequations}
\begin{align}
    \bm{G}_\alpha &= {\rm diag}\{ \alpha \bm{E}, 1 \}, \ \mbox{and}
    \\
    \bm{A}_\alpha
    &=
    \mathrm{diag}
    \{
    \bm{E}, \ \alpha\bm{E}, \ 1
    \} .
\end{align}
\end{subequations}
We can easily \replaced{find the following relationship:}{obtain $\bm{G}_{\rm SIM} = \bm{G}_{\rm SE} \bm{G}_{\alpha}$.}
\begin{align}
    \label{eq:sim3_as_semidirect_product}
    \bm{G}_{\rm SIM} = \bm{G}_{\rm SE} \bm{G}_{\alpha}.
\end{align}
Note that this multiplication is not commute.
\added{The set of $\bm{G}_{\alpha}$ is equivalent to $\mathbb{R}^{+} := \{ \alpha \in \mathbb{R} | \alpha > 0\}$.
Equation \eqref{eq:sim3_as_semidirect_product} represents that SIM(3) is a semidirect product of SE(3) and $\mathbb{R}^{+}$, which is an important property used for deriving dynamics of the PC7 model in Section \ref{sect:microstretch_rod}.}
\deleted{Adjoint representation of $\bm{G}_\alpha$ is given as}

\section{Poincar\'e Equations for Cosserat Rod \cite{boyer2017poincare}\label{sect:cosserat_rod}}
\subsection{Configuration of Cosserat Rod}
The state of a continuum is represented by a configuration, which is determined by a smooth mapping from the material space to the geometric space.
The material space is a time-independent parameter space fixed to each infinitesimal element of the continuum, whereas the geometric space is the actual \replaced{3D}{three-dimensional} Euclidean space in which physical phenomena occur.
The configuration at time $t=0$, where the continuum is in an energetically stationary and stable state, is called the reference configuration.
The configuration at an arbitrary time $t>0$, which has changed from the reference configuration, is called the deformed configuration.

A classical continuum is represented as a collection of material points on the spatial coordinates of the \replaced{3D}{three-dimensional} material space.
In contrast, a Cosserat continuum is represented as a collection of microstructures placed at each point of the continuum.
In particular, a Cosserat continuum mathematically considers \replaced{3D}{three-dimensional} rigid microstructures \deleted{(rigid frames possessing independent rotational and translational degrees of freedom)} arranged along an $n \leq 3$ dimensional continuum. From a physical modeling perspective, this formulation is \replaced{suitable}{highly suited} for dimensionally reduced models (such as beams or shells), where the remaining $(3-n)$ dimensions—representing cross-sections or thickness—are treated as rigid microstructures. A rigid microstructure is a microscopic rigid body that does not allow microscopic deformation.
\deleted{(次のセクションで説明されているので、ここでは省略) In a microstretch continuum \cite{eringen2012microcontinuum}, a stretch microstructure is introduced by extending the rigid microstructure so that it allows isotropic expansion and contraction in addition to rigid rotation and translation.}

In this study, we consider \replaced{a Cosserat rod, which is rod-shaped Cosserat continuum in the case of $n=1$}{the rod-shaped Cosserat continuum corresponding to the case $n=1$, namely, a Cosserat rod}.
For a Cosserat rod, the centerline of the rod is regarded as the material space of the \replaced{1D}{one-dimensional} continuum and is parameterized by the spatial variable $s$.
At time $t$, the rigid microstructure attached to each point $s$ can be described in the geometric space by a rotation $\bm{R}(t,s)\in \mathrm{SO}(3)$ and a position $\bm{p}(t,s)\in \mathbb{R}^3$. 
Hereafter, we represent the configuration using $\bm{G}_{\rm SE}(t,s)\in \mathrm{SE}(3)$, \deleted{which combines the orientation $\bm{R}$ and the position $\bm{p}$,} defined as
\begin{align}
    \bm{G}_{\rm{SE}}(t,s) = 
    \bm{G}_{\rm{SE}}(\bm{R}(t,s), \bm{p}(t,s)).
\end{align}

\subsection{Lagrangian Density}
In a standard rigid-link system, the configuration is described by a finite-dimensional generalized coordinate $\bm{q}(t)$ that depends only on time. 
Accordingly, the Lagrangian is defined as a function of the configuration and its time derivative, typically written as $L(\bm{q},\dot{\bm{q}})$, and Hamilton's principle considers variations over time.
\replaced{However}{In contrast}, the configuration of a Cosserat rod is distributed continuously along the material coordinate $s$.
Therefore, the configuration is represented as a field $\bm{G}_{\rm SE}(t,s)$ that depends on both time $t$ and space $s$.
Because the rod has infinite \replaced{DOFs}{degrees of freedom} along the spatial direction, its energy cannot be described only by a finite-dimensional Lagrangian\replaced{, but by spatial integral of {\it Lagrangian density}}{. 
Instead, we introduce the Lagrangian density} \replaced{$\mathcal{L}$}{$\mathfrak{L}$}\replaced{ that}{, which} represents the energy per unit length of the rod in the material space\replaced{ as follows:}{. This density $\mathfrak{L}$ depends on the configuration field $\bm{G}_{\rm{SE}}$, its temporal derivative $\dot{\bm{G}}_{\rm SE}$, and its spatial derivative $\bm{G}_{\rm SE}^\prime$; therefore, we can write it as $\mathfrak{L}(\bm{G}_{\rm{SE}},\dot{\bm{G}}_{\rm{SE}},\bm{G}_{\rm{SE}}')$. }

\deleted{The total Lagrangian of the Cosserat rod is then obtained by integrating the Lagrangian density over the material space of the rod as}
\begin{align}
 L(t) = \int_s \mathcal{L}(\bm{G}_{\rm{SE}},\dot{\bm{G}}_{\rm{SE}},\bm{G}_{\rm{SE}}') ds. \label{eq:lagrangian density}
\end{align}

\added{As written in \eqref{eq:lagrangian density}, $\mathcal{L}$ depends on the configuration field $\bm{G}_{\rm{SE}}$, its temporal derivative $\dot{\bm{G}}_{\rm SE}$, and its spatial derivative $\bm{G}_{\rm SE}^\prime$.}
Exploiting the left-invariance under superposed rigid transformations, \deleted{the Lagrangian density} $\mathcal{L}$ can be reduced to a form that does not explicitly depend on $\bm{G}_{\rm SE}$ \added{as follows} \cite{boyer2017poincare}. 
\replaced{We introduce}{By introducing} the left-invariant vector fields in the Lie algebra $\mathfrak{se}(3)$ defined \replaced{so as to satisfy}{as}
\begin{align}
\label{eq:se3_tangent_vectors}
    [\bm{\mathcal{V}}_{\mathfrak{se}}\times_4] = \bm{G}_{\mathrm{SE}}^{-1}\dot{\bm{G}}_{\mathrm{SE}}
    \ \mbox{and} \
    [\bm{\xi}_{\mathfrak{se}}\times_4] = \bm{G}_{\mathrm{SE}}^{-1}\bm{G}_{\mathrm{SE}}'
    .
\end{align}
\added{The 6D vectors $\bm{\mathcal{V}}_{\mathfrak{se}}$ and $\bm{\xi}_{\mathfrak{se}}$ can be calculated as follows:}
\begin{subequations}
\begin{align}
    \bm{\mathcal{V}}_{\mathfrak{se}}
    &= 
    \begin{bmatrix}
        (\bm{R}^T \bm{\omega})^T & (\bm{R}^T \dot{\bm{p}})^T 
    \end{bmatrix}^T, \ \mbox{and}
    \\
    \bm{\xi}_{\mathfrak{se}}
    &= 
    \begin{bmatrix}
        (\bm{R}^T \bm{u})^T & (\bm{R}^T \bm{p}^\prime)^T 
    \end{bmatrix}^T
\end{align}
\end{subequations}
where $\bm{\omega}$ and $\bm{u}$ are 3D vectors that represent Lie algebra of $SO(3)$:
\begin{align}
    [\bm{\omega} \times_3] = \dot{\bm{R}} \bm{R}^T,
    \ \mbox{and} \
    [\bm{u} \times_3] = \bm{R}^\prime \bm{R}^T.
\end{align}
\added{$\bm{\mathcal{V}}_{\mathfrak{se}}$ is well-known spatial velocity that is often used in robotics.
In \cite{boyer2017poincare}, $\bm{\xi}_{\mathfrak{se}}$ was referred to as {\it strain} vector.}

\added{Finally, the Lagrangian density is left-invariant to any arbitrary rigid transformation $\bm{H}_{\rm SE} \in$ SE(3). 
Namely, the value of $\mathcal{L}$ remains the same even when $\bm{G}_{\rm SE}$, $\dot{\bm{G}}_{\rm SE}$ and $\bm{G}_{\rm SE}^{\prime}$ are transformed by left multiplication with $\bm{H}_{\rm{SE}}$:}
\begin{align}
    \label{eq:left_invariant_se3}
    \mathcal{L}(\bm{G}_{\rm{SE}},\dot{\bm{G}}_{\rm{SE}},\bm{G}_{\rm{SE}}')
    =
    \mathcal{L}(\bm{H}_{\rm SE}\bm{G}_{\rm{SE}}, \bm{H}_{\rm SE}\dot{\bm{G}}_{\rm{SE}}, \bm{H}_{\rm SE}\bm{G}_{\rm{SE}}') .
\end{align}
\added{Choosing $\bm{H}_{\rm SE} = \bm{G}_{\rm{SE}}^{-1}$, we can reduce }the Lagrangian density \deleted{is reduced} as follows:
\begin{align}
    \label{eq:lagrangian}
\mathcal{L}(\bm{G}_{\rm{SE}},\dot{\bm{G}}_{\rm{SE}},\bm{G}_{\rm{SE}}')
=
\mbox{\replaced{$\mathcal{L}_{\frak{se}}$}{$\mathcal{L}$}}(\bm{\mathcal{V}}_{\mathfrak{se}},\bm{\xi}_{\mathfrak{se}}).
\end{align}

\subsection{Derivation of Poincar\'e Equations}
In classical rigid-body dynamics, the equations of motion are derived from the Lagrangian by applying the extended Hamilton's principle, where the variation is taken \replaced{w.r.t.}{with respect to} time. 
For a Cosserat rod, the \deleted{configuration is distributed along the spatial coordinate $s$, and therefore the extended Hamilton's principle is applied to the Lagrangian by considering variations with respect to both time and space. 
The} extended Hamilton's principle for the total Lagrangian $L(t)$ of the rod is written \replaced{using the following temporal- and spatial-integral:}{as}
\begin{align}
  \delta \int_t \int_s \mathcal{L} ds dt &= -\int_t \delta W_{\rm ext} dt\label{eq:hamilton}
\end{align}
where $\delta$ denotes a variation with the coordinates $(t,s)$ fixed. 
\added{$\delta W_{\rm ext}$ is the virtual work of external force $\bm{\mathcal{F}}_{\rm ext}$, represented as}
\begin{align}
  \delta W_{\rm ext} = \int_s \bm{\zeta}_{\frak{se}}^T \bm{\mathcal{F}}_{\rm ext} ds 
  \label{eq:virtual work}
\end{align}
\replaced{where $\bm{\zeta}_{\mathfrak{se}}$ is }{We introduce the virtual displacement represented by} the left-invariant vector field \replaced{defined}{$\bm{\zeta}_{\mathfrak{se}}$} as
\begin{align}
  [\bm{\zeta}_{\frak{se}}\times_4] = \bm{G}_{\rm{SE}}^{-1}\delta \bm{G}_{\rm{SE}} \in \mathfrak{se}(3). \label{eq:zeta}
\end{align}
\deleted{Using this virtual displacement, the virtual work of the external force is given by}
\deleted{where $\bm{\mathcal{F}}_{\rm ext}$ denotes the generalized external force.}
The variations of $\bm{\mathcal{V}}_{\mathfrak{se}}$ and $\bm{\xi}_{\mathfrak{se}}$ are obtained by taking the variations of both sides of (\ref{eq:se3_tangent_vectors}) and using (\ref{eq:zeta})\replaced{, which can be}{. 
They are} expressed as
\begin{subequations}
\label{eq:delta_V}
\begin{align}
  \delta \bm{\mathcal{V}}_{\frak{se}} &= \dot{\bm{\zeta}}_{\frak{se}} + [\bm{\mathcal{V}}_{\frak{se}}\times_6]\bm{\zeta},
  \ \mbox{\added{and}} 
  \\
  \delta \bm{\xi}_{\frak{se}} &= \bm{\zeta}_{\frak{se}}' + [\bm{\xi}_{\frak{se}}\times_6]\bm{\zeta}.
\end{align}    
\end{subequations}
Applying the standard variational calculus with integration by parts in time and space, the Poincar\'e equations for the Cosserat rod are derived as
\begin{align}
  &\dfrac{\partial}{\partial t}\left(\dfrac{\partial\mathcal{L}}{\partial\bm{\mathcal{V}}_{\frak{se}}}\right) - [\bm{\mathcal{V}}_{\frak{se}}\times_6]^T\left(\dfrac{\partial\mathcal{L}}{\partial\bm{\mathcal{V}}_{\frak{se}}}\right)\notag\\
  &+ \dfrac{\partial}{\partial s}\left(\dfrac{\partial\mathcal{L}}{\partial\bm{\xi}_{\frak{se}}}\right) - [\bm{\xi}_{\frak{se}}\times_6]^T\left(\dfrac{\partial\mathcal{L}}{\partial\bm{\xi}_{\frak{se}}}\right) = \bm{\mathcal{F}}_{\rm ext}. \label{eq:poincare_cosserat}
\end{align}
\replaced{This equation represents}{These equations represent} the reduced dynamics of the Cosserat rod, expressed simply in terms of the local velocity and strain within the Lie algebra $\mathfrak{se}(3)$ \cite{boyer2017poincare}.

\section{Poincar\'e Equations for Microstretch Rod\label{sect:microstretch_rod}}
\subsection{Configuration of Microstretch Rod and its Lagrangian Density}
A microstretch \cite{eringen2012microcontinuum} rod is an extension of the Cosserat rod in which the rigid microstructure attached to each point $s$ \replaced{can isotropically extract and contract}{is allowed to undergo isotropic expansion and contraction}. \replaced{Hereafter}{In what follows, to distinguish it from the rigid microstructure}, we refer to this extended microstructure as a \replaced{stretch microstructure to distinguish it from the rigid microstructure}{stretch microstructure}.
The configuration of the stretch microstructure is described not only by \replaced{$\bm{G}_{\rm SE}$}{the elements of a rigid transformation, namely the orientation $\bm{R}$ and the position $\bm{p}$,} but also by a stretch ratio \added{(or so-called {\it scale})} $\alpha\in \mathbb{R}^{+}$ representing the isotropic expansion. 
We define the reference configuration \deleted{by} setting the stretch ratio as $\alpha=1$. 
The configuration of the stretch microstructure in the geometric space can then be represented by an element $\bm{G}_{\rm{SIM}}$ \deleted{$\in \mathrm{SIM}(3)$} of the similarity transformation group \added{$\mathrm{SIM}(3)$}.

\replaced{T}{Similarly to the rigid microstructure, t}he Lagrangian density $\mathcal{L}$ for the stretch microstructure can be \replaced{calculated}{defined} in \replaced{a manner similar to \eqref{eq:lagrangian density} but}{the same form as Eq.~(\ref{eq:lagrangian density}) by} using $\bm{G}_{\rm{SIM}}$ instead of $\bm{G}_{\rm{SE}}$ \added{: $\mathcal{L}(\bm{G}_{\rm{SIM}},\dot{\bm{G}}_{\rm{SIM}},\bm{G}^{\prime}_{\rm{SIM}})$}.
\added{From the left-invariant property, the value of the Lagrangian density is invariant under a rigid transformation $\bm{H}_{\rm{SE}}\in \mathrm{SE}(3)$, in a manner similar to \eqref{eq:left_invariant_se3}:}
\begin{align}
    \label{eq:left_invariant_sim3}
    & \mathcal{L}(\bm{G}_{\rm{SIM}},\dot{\bm{G}}_{\rm{SIM}},\bm{G}^{\prime}_{\rm{SIM}})
    \nonumber
    \\
    & \hspace{3em}
    = 
    \mathcal{L}(\bm{H}_{\rm{SE}}\bm{G}_{\rm{SIM}}, \bm{H}_{\rm{SE}}\dot{\bm{G}}_{\rm{SIM}}, \bm{H}_{\rm{SE}}\bm{G}^{\prime}_{\rm{SIM}}).
\end{align}
\added{Choosing $\bm{H}_{\rm{SE}} = \bm{G}_{\rm{SE}}^{-1}$ and substituting \eqref{eq:sim3_as_semidirect_product}, we can calculate the first argument in the right-hand side of \eqref{eq:left_invariant_sim3} as}
\begin{align}
    \label{eq:sim3_reduction01}
    \bm{H}_{\rm{SE}}\bm{G}_{\rm{SIM}} 
    =
    \bm{G}_{\rm{SE}}^{-1} \bm{G}_{\rm{SE}} \bm{G}_{\alpha}
    =
    \bm{G}_{\alpha} .
\end{align}
\deleted{Since the Lagrangian density is invariant under rigid transformations, the value of $\mathcal{L}$ remains the same even if $\bm{G}_{\rm{SIM}}\in \mathrm{SIM}(3)$ is transformed into $\bm{H}_{\rm{SE}}\bm{G}_{\rm{SIM}}$ by left multiplication with an arbitrary constant rigid transformation $\bm{H}_{\rm{SE}}\in \mathrm{SE}(3)$.
By choosing $\bm{H}_{\rm{SE}}=\bm{G}_{\rm{SE}}^{-1}$, the Lagrangian density $\mathcal{L}(\bm{G}_{\rm{SIM}},\dot{\bm{G}}_{\rm{SIM}},\bm{G}_{\rm{SIM}}')$ can be reduced to a form that does not explicitly depend on $\bm{G}_{\rm{SE}}$, as follows:}
\replaced{The second and third arguments can be expressed using $\bm{\mathcal{V}}_{\frak{sim}}$ and $\bm{\xi}_{\frak{sim}}$, which are}{By expressing the above equation in terms of} the left-invariant vector fields on $\mathrm{SIM}(3)$\deleted{, as} defined \replaced{as follows:}{in Eqs.~(\ref{eq:V_sim3}) and (\ref{eq:xi_sim3}), we obtain}
\begin{subequations}
\label{eq:V_sim3} 
\begin{align}
  [\bm{\mathcal{V}}_{\frak{sim}}\times_4] = \bm{G}_{\rm SIM}^{-1}\dot{\bm{G}}_{\rm SIM},
  \\
  [\bm{\xi}_{\frak{sim}}\times_4] = \bm{G}^{-1}_{\rm SIM}\bm{G}_{\rm SIM}^{\prime} .
\end{align}    
\end{subequations}
\replaced{These $\bm{\mathcal{V}}_{\frak{sim}}$ and $\bm{\xi}_{\frak{sim}}$ are 7D vectors calculated as follows:}{Since $\bm{\mathcal{V}}_{\frak{sim}}$ and $\bm{\xi}_{\frak{sim}}$ can be expressed using $\bm{\mathcal{V}}_{\frak{se}}=[\bm{\mathcal{V}}^T_{\frak{se}a} \ \bm{\mathcal{V}}^T_{\frak{se}l}]^T$
and
$\bm{\xi}_{\frak{se}}= [\bm{\xi}^T_{\frak{se}a} \
    \bm{\xi}^T_{\frak{se}l}]^T$
as follows,}
\begin{subequations}
\begin{align}
  \bm{\mathcal{V}}_{\frak{sim}} &= \begin{bmatrix}
    (\bm{R}^T \bm{\omega})^T &
    \alpha^{-1} (\bm{R}^T \dot{\bm{p}})^T &
    \alpha^{-1}\dot{\alpha}
  \end{bmatrix}^T\label{eq:V_sim3_2}
  \\
  \bm{\xi}_{\frak{sim}} &= \begin{bmatrix}
    (\bm{R}^T \bm{u})^T &
    \alpha^{-1} (\bm{R}^T \bm{p}^\prime)^T &
    \alpha^{-1}\alpha^\prime
  \end{bmatrix}^T. \label{eq:xi_sim3_2}
\end{align}
\end{subequations}
\added{Using $\bm{\mathcal{V}}_{\frak{sim}}$ and $\bm{\xi}_{\frak{sim}}$, we can calculate the second and third arguments in the right-hand side of \eqref{eq:left_invariant_sim3} as follows:}
\begin{subequations}
    \label{eq:sim3_reduction02}
\begin{align}
    \bm{H}_{\rm SE} \dot{\bm{G}}_{\rm SIM}
    &=
    \bm{G}_{\alpha} [\bm{\mathcal{V}}_{\frak{sim}} \times_4 ],
    \ \mbox{and} 
    \\
    \bm{H}_{\rm SE} \bm{G}_{\rm SIM}^{\prime}
    &=
    \bm{G}_{\alpha} [\bm{\xi}_{\frak{sim}} \times_4 ]
\end{align}    
\end{subequations}

\replaced{Finally, from \eqref{eq:sim3_reduction01} and \eqref{eq:sim3_reduction02}}{Thus}, the Lagrangian density can be rewritten as a function of $\bm{\mathcal{V}}_{\frak{sim}}$, $\bm{\xi}_{\frak{sim}}$, and $\alpha$:
\begin{align}
    \mathcal{L}(\bm{G}_{\rm{SIM}},\dot{\bm{G}}_{\rm{SIM}},\bm{G}^{\prime}_{\rm{SIM}})
  =
  \mbox{\replaced{$\mathcal{L}_{\frak{sim}}$}{$\mathcal{L}$}}(\bm{\mathcal{V}}_{\frak{sim}},\bm{\xi}_{\frak{sim}},\alpha).\label{eq:lagrangian(microstretch)}
\end{align}
By introducing the left-invariant vector fields on SIM(3), the velocity and spatial derivatives on the group\added{, $\bm{\mathcal{V}}_{\frak{sim}}$ and $\bm{\xi}_{\frak{sim}}$,} can be expressed in the material frame.
Therefore, \deleted{compared with \eqref{eq:lagrangian(microstretchSE3)},} \added{$\mathcal{L}_{\frak{sim}}$ in} \eqref{eq:lagrangian(microstretch)} \deleted{can be written as a function that} does not explicitly include $\dot{\alpha}$ and $\alpha'$ as independent variables.
\replaced{However}{On the other hand}, this reduction does not eliminate the scale $\alpha$ itself, which stems from $\bm{G}_\alpha$ in \eqref{eq:sim3_reduction01} and \eqref{eq:sim3_reduction02}.
Since the Lagrangian density is not invariant under scale transformations, \replaced{$\mathcal{L}_{\frak{sim}}$}{$\mathcal{L}$ in \eqref{eq:lagrangian(microstretch)}} still depends explicitly on \deleted{the scale} $\alpha$ after reduction.

In \replaced{the previous section}{\eqref{eq:lagrangian}}, the components $\bm{R}$ and $\bm{p}$ of $\bm{G}_{\rm SE}$ \deleted{represent the rotational attitude and position of the rigid microstructure, respectively.
These quantities} depend on the choice of the reference coordinate frame\replaced{, which allows us to remove this dependence}{, and this dependence can be removed by} acting from the left with $\bm{G}_{\rm SE}^{-1}$ \added{in \eqref{eq:lagrangian}}.
\replaced{However}{In contrast}, the scale $\alpha$ is an internal variable representing the expansion ratio of the stretch microstructure itself, and \replaced{does not depend on}{it is not changed by} the choice of the reference coordinate frame.
Therefore, $\alpha$ \replaced{cannot be}{is not a variable} eliminated by Lagrangian reduction, but is instead treated as a parameter in the reduced Lagrangian density and the subsequent Poincar\'e equations.

\subsection{Derivation of Poincar\'e Equations for Microstretch Rod}
In \replaced{this subsection}{what follows}, we apply the extended Hamilton principle \replaced{using $\mathcal{L}_{\frak{sim}}$ in}{to} \eqref{eq:lagrangian(microstretch)}, in \replaced{a manner similar to}{the same manner as in} the previous section.
Hereafter, we \replaced{use the notation such as}{simply denote $\bm{G}_{\rm{SIM}}$, $\bm{\mathcal{V}}_{\frak{sim}}$, and $\bm{\xi}_{\frak{sim}}$ by} $\bm{G}$, $\bm{\mathcal{V}}$, and $\bm{\xi}$, \replaced{omitting the index from $\bm{G}_{\rm{SIM}}$, $\bm{\mathcal{V}}_{\frak{sim}}$, and $\bm{\xi}_{\frak{sim}}$ for simplicity}{respectively}.
\replaced{First, the}{The} virtual displacement $\bm{\zeta}$ \replaced{can be defined in a manner similar to \eqref{eq:zeta} as}{and the variations of $\bm{\mathcal{V}}$, $\bm{\xi}$, and $\alpha$ are expressed as follows:}
\begin{align}
  [\bm{\zeta}\times_4] &= \bm{G}^{-1}\delta \bm{G} \in \mathfrak{sim}(3). \label{eq:virtual displacement}
\end{align}
\added{Then, the variations of $\bm{\mathcal{V}}$, $\bm{\xi}$, and $\alpha$ are expressed as follows:}
\begin{align}
  \delta \bm{\mathcal{V}} = \dot{\bm{\zeta}} + [\bm{\mathcal{V}}\times_7]\bm{\zeta},
  \
  \delta \bm{\xi} = \bm{\zeta}' + [\bm{\xi}\times_7]\bm{\zeta},
  \
  \delta\alpha 
  = \alpha\bm{\zeta}^T\bm{e}_7.
\end{align}
\replaced{Finally}{From the above relations}, the Poincar\'e equation\deleted{s} for the microstretch rod can be derived \replaced{in a manner similar to that of}{the same manner as for} the Cosserat rod in \replaced{\eqref{eq:poincare_cosserat}}{Eq.~(\ref{eq:poincare_cosserat})} as follows:
\begin{align}
  &\pdif{}{t}
  \left(\dfrac{\partial \mathcal{L}}{\partial \bm{\mathcal{V}}}\right) - [\bm{\mathcal{V}}\times_7]^T \dfrac{\partial \mathcal{L}}{\partial \bm{\mathcal{V}}}  
  \nonumber
  \\
  &+\dfrac{\partial}{\partial s}\left(\dfrac{\partial \mathcal{L}}{\partial \bm{\xi}}\right) - [\bm{\xi}\times_7]^T \dfrac{\partial \mathcal{L}}{\partial \bm{\xi}}
    -\alpha \dfrac{\partial \mathcal{L}}{\partial \alpha}\bm{e}_7
  = \bm{\mathcal{F}}_{\rm ext}. \label{eq:poincare}
\end{align}

\replaced{The structure of \eqref{eq:lagrangian(microstretch)}}{This structure} is consistent with \replaced{that in the literature}{the formulation by Holm et al.} \cite{holm1998euler} for Euler--Poincar\'e reduction with a parameter-dependent Lagrangian.
\deleted{In the present formulation, the scale $\alpha$ is not a velocity variable eliminated by left-invariant reduction, but appears as a parameter that remains in the reduced Lagrangian density.}
\replaced{The}{Consequently, the} variation \replaced{w.r.t.}{with respect to} $\alpha$ \replaced{stems}{corresponds to an additional term arising} from the semidirect-product structure \added{given by \eqref{eq:sim3_as_semidirect_product}}, and \replaced{results to}{gives} \replaced{a formulation similar to \eqref{eq:poincare_cosserat}, but with an additional term in the left-hand side of \eqref{eq:poincare},}{the term}
$-\alpha(\partial \mathcal{L}/\partial \alpha)\bm{e}_7$\deleted{ in the Poincar\'e equation for the stretch microstructure}.

\replaced{Although}{While} a recent study \deleted{by Tobin et al.}\cite{Tobin2025} introduced additional \replaced{DOFs}{degrees of freedom} to represent complex cross-sectional deformations, their formulation \replaced{did not follow}{deviates from} the strict Lie group framework, \replaced{prohibiting us from applying}{making} the Euler--Poincar\'e reduction mathematically\deleted{ prohibitive}. 
\replaced{Our approach focuses}{In our approach, by focusing} specifically on the isotropic expanding and contracting deformations \replaced{and reduces}{typical of artificial muscles, we successfully reduce} the Lagrangian density to a form depending solely on the Lie algebra $\mathfrak{sim}(3)$ and the scaling parameter $\alpha$. This preservation of the algebraic structure \replaced{enables us to apply}{is precisely what enables the seamless application of} the formulation by \deleted{Holm et al.}\cite{holm1998euler} for Euler--Poincar\'e reduction with a parameter-dependent Lagrangian.

\begin{figure}[tbp]
  \begin{center}
    \includegraphics[width=0.6\hsize]{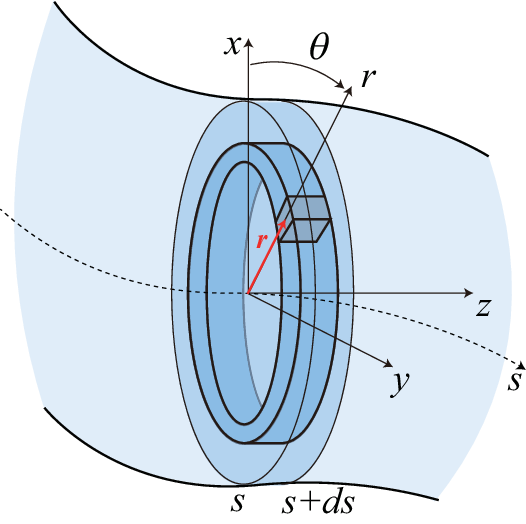}
  \end{center}
  \caption{Cross section of the stretch miscrostructure in a microstretch rod.
  Without loss of generality, the $z$-axis of the local frame can be set aligned to the tangent direction of $s$.}
  \label{fig:pc7_model_microstructure}
\end{figure}
\subsection{Detailed Formulation of Lagrangian Density}
\added{In this subsection, we provide more explicit formulation of the Lagrangian density for the next subsection.
Hereafter, we assume an axisymmetric structure for simplicity. 
The Lagrangian density $\mathcal{L}$ consists of the kinetic energy density $\mathcal{K}$ and the potential energy density $\mathcal{U}$ as follows:
\begin{align}
    \mathcal{L} = \mathcal{K} - \mathcal{U}.
\end{align}
In the rest of this subsection, we formulate $\mathcal{K}$ and $\mathcal{U}$ in detail.}

\added{Let us consider the cross section of the stretch miscrostructure in a microstretch rod, 
as shown in \figref{fig:pc7_model_microstructure}.
An infinitesimal volume element the material space $s$ can be visualized as a {\it ring} shown in \figref{fig:pc7_model_microstructure}.
Its volume $dV_0$ is calculated as}
\begin{align}
    dV_0 = r d\theta dr \mbox{\added{$ds$}}
\end{align}
\added{where $(r, \theta)$ is the polar coordinate set in the cross section.}
\replaced{Because this paper focuses on}{Considering} the isotropic expansion and contraction of the microstretch rod, the cross-section scales uniformly by the parameter $\alpha$\deleted{, while the axial deformation is represented by $\alpha \xi_6$}. 
Therefore, the circle length, thickness and width of the ring change from $r d\theta, dr, ds$ to $\alpha r d\theta, \alpha dr, \alpha \bm{\xi}^T\bm{e}_6 ds$ by deformation, respectively.
\added{The resultant volume $dV$ in the deformed configuration can be calculated as}
\begin{subequations}
\label{eq:volume_change}
\begin{align}
    dV 
    = (\alpha r d\theta)(\alpha dr)(\alpha \xi_6 ds)
    = v dV_0,
    \\
    \xi_6 := \bm{\xi}^T\bm{e}_6
    \\
    v := \alpha^3 \xi_6
\end{align}
\end{subequations}
\replaced{where $v$ can be regarded as volumetric scale factor}{This relation indicates that the volumetric scale factor associated with the deformation is given by $\alpha^3 \xi_6$}.
\added{Importantly, we assume the mass conservation law before and after deformation as follows:}
\begin{align}
    \label{eq:mass_conservation}
    \rho(s,t) dV = \rho_0(s) dV_0
\end{align}
\added{where $\rho_0(s)$ and $\rho(s,t)$ are the density before and after deformation, respectively.
From \eqref{eq:volume_change} and \eqref{eq:mass_conservation}, we obtain}
\begin{align}
    \rho_0 (s) = \rho(s,t) v .
\end{align}

\deleted{Hereafter, to simplify the formulation, we assume an axisymmetric structure. 
The Lagrangian density $\mathcal{L}$ is defined as the difference between the kinetic energy density $\mathcal{K}$ and the potential energy density $\mathcal{U}$, namely, $\mathcal{L} = \mathcal{K} - \mathcal{U}$. }

\added{Without loss of generality, the $z$-axis of the local frame can be set aligned to the tangent direction of $s$, as shown in Fig. \ref{fig:pc7_model_microstructure}.}
\replaced{The local position of an infinitesimal element is denoted by}{A vector $\bm{r}$ on the cross-section at the material coordinate $s$ is defined using the dependent spatial variables $r(s)$ and $\theta(s)$ as follows:}
\begin{align}
    \bm{r}_i(r,\theta) = r
    \begin{bmatrix}
        \cos\theta & \sin\theta & 0
    \end{bmatrix}^T .
\end{align}
\added{Then, the 3D linear velocity in the deformed configuration is obtaned as}
\begin{subequations}
\begin{align}
    \bm{v}_i
    =
    \alpha & \bm{B}_i(r, \theta) \bm{\mathcal{V}},
    \\
    & \bm{B}_i(r, \theta)
    :=
    \begin{bmatrix}
        -[\bm{r}_i \times_3] & \bm{E} & \bm{r}_i
    \end{bmatrix}.
\end{align}
\end{subequations}
\deleted{Let $\rho(t,s)$ denote the mass density of the rod, and let $\bm{v}_p$ denote the spatial velocity of a point $\bm{r}$ on the cross-section in the reference configuration, given by $\bm{v}_p = (\mathcal{V}_{\lambda}+[\bm{\mathcal{V}}_{a}\times_3])\alpha \bm{r} + \alpha \bm{\mathcal{V}}_{l}$. }
\subsubsection{Kinetic Energy Density} \replaced{The total kinetic energy $K$ of the miscrostretch rod is calculated as}{$\mathcal{K}$ is then calculated by integrating the kinetic energy of the infinitesimal mass elements over the cross-section:}
\begin{align}
  \label{eq:kinetic_energy}
  K 
  =
  \frac{1}{2}
  \int_s \int_{r,\theta}
  \rho\bm{v}^T_i \bm{v}_i dV
  =
  \frac{1}{2}
  \int_s \int_{r,\theta}
  \rho_0 \bm{v}^T_i \bm{v}_i dV_0 .
\end{align}
\added{We used the mass conservation law \eqref{eq:mass_conservation} in the above transformation.
Subsequently, we can derive the kinetic energy density $\mathcal{K} = \pdif{K}{s}$ as}
\begin{subequations}
\begin{align}
  \mathcal{K} 
  =
  \dfrac{1}{2} \bm{\mathcal{V}}^T 
  & \mbox{\replaced{$\widehat{\bm{\mathcal{M}}}$}{$\alpha ^2 \bm{\mathcal{M}}$}} 
  \bm{\mathcal{V}}, \label{eq:mathcal{K}}
  \\
  \widehat{\bm{\mathcal{M}}} &:= \alpha ^2 \bm{\mathcal{M}}_0,
  \\
  \mbox{\replaced{$\bm{\mathcal{M}}_0$}{$\bm{\mathcal{I}}_0$}}
  &:=
  \rho_0 \int_{r, \theta} \bm{B}^T \bm{B} r dr d\theta
  \nonumber
  \\
  &=
  \mathrm{diag}\{ J_{x0},J_{y0},J_{z0},A_0,A_0,A_0,J_{z0} \}
\end{align}
\end{subequations}
\replaced{where}{The matrix $\bm{\mathcal{I}}_0(s)$ is expressed using the second moments of area $J_{x0}(s)$ and $J_{y0}(s)$ about the $x$- and $y$-axes of the stretch microstructure, the polar second moment of area $J_{z0}(s)$ about the $z$-axis, and the cross-sectional area $A_0(s)$ in the reference configuration.}
\begin{itemize}
    \item \added{$\bm{\mathcal{M}}_0$ is the time-invariant inertia matrix expressed in the reference configuration, consisting of the second moments of area $J_{x0}(s)$ and $J_{y0}(s)$ about the $x$- and $y$-axes of the stretch microstructure, the polar second moment of area $J_{z0}(s)$ about the $z$-axis, and the cross-sectional area $A_0(s)$.}
    \item \added{$\widehat{\bm{\mathcal{M}}}$ can be regarded as {\it scaled} inertia matrix, obtained as $\bm{\mathcal{M}_0}$ multiplied with $\alpha^2$.}
\end{itemize}

\subsubsection{Potential Energy Density}
\added{The potential energy density $\mathcal{U}$ is calculated by integrating the elastic energy over the cross-section.}
\deleted{The strain field $\bm{\xi}(s)$ in the reference configuration is set as $\bm{\xi}_0(s)=\bm{e}_6$.\textcolor{red}{(この設定(仮定)は必ずしも必要ないのでは?)}}
\deleted{Similarly, the potential energy density $\mathcal{U}$ is calculated by integrating the elastic energy over the cross-section.} 
\replaced{We can calculate the 3D local strain vector $\bm{\epsilon}_i$ of the infinitesimal mass element as}{Let $E$ and $G$ denote the Young's modulus and the shear modulus, respectively. To simplify the notation, we introduce the local strain vector $\bm{\epsilon}_p = (\xi_{\lambda}+[\bm{\xi}_a\times_3])\alpha \bm{r} + \alpha \bm{\xi}_{l} - \bm{\xi}_{0l}$. }
\begin{align}
    \bm{\epsilon}_i = \alpha \bm{B}_i (r, \theta) \varDelta \bm{\xi},
    \quad
    \varDelta \bm{\xi} := \bm{\xi} - \alpha^{-1}\bm{\xi}_0
\end{align}
\added{where $\bm{\xi}_0$ is the 7D strain vector in the reference configuration.}
\replaced{Using this strain vector, we can formulate the total potential energy}{Thus, $\mathcal{U}$ is formulated} as follows:
\begin{align}
  \label{eq:potential_energy_density_derivation}
  U 
  =
  &\dfrac{1}{2} \int_s \int_{r, \theta}
  \left\{ 
  \bm{\epsilon}_i^T \mathrm{diag}(G,G,E) \bm{\epsilon}_i \right.\notag\\
  & \hspace{5em} +
  \left.
  E (\alpha - 1)^2
  +
  E (\alpha - 1)^2
  \right\} dV
  \nonumber
  \\
  =
  &\dfrac{1}{2} \int_s \int_{r, \theta}
  v \left\{ 
  \bm{\epsilon}_i^T \mathrm{diag}(G,G,E) \bm{\epsilon}_i 
  +
  2E (\alpha - 1)^2
  \right\} dV_0
\end{align}
\added{where $E$ and $G$ are the Young's modulus and the shear modulus, respectively.
The potential energy consists of three terms. 
The first term represents the elastic energy associated with extension, shear, and bending along the $s$-axis. 
The second term accounts for the energy resulting from the change in the circumferential length of the infinitesimal element from $r d\theta$ to $\alpha r d\theta$.
The third term corresponds to the energy due to the change in the radial length of the infinitesimal element from $dr$ to $\alpha dr$.
Note that we exclude the gravity term here but consider it as an external force later.}

\added{We can calculate the potential energy density $\mathcal{U} = \pdif{U}{s}$ from \eqref{eq:potential_energy_density_derivation} as follows:}
\begin{subequations}
\begin{align}
  \mathcal{U} &=
  \dfrac{1}{2} \varDelta \bm{\xi}^T \widehat{\bm{\Sigma}} \varDelta \bm{\xi}
  +
  E \mbox{\replaced{$v$}{$\alpha^3 \xi_6$}} A_0 (\alpha - 1)^2.\label{eq:mathcal{U}}
  \\
  &\widehat{\bm{\Sigma}} := \alpha^2 \bm{\Sigma}_0
  \\
  &\bm{\Sigma}_0
  :=
  E~\mathrm{diag}\{ 1, 1, u, u, u, 1, u \}
  \mbox{\deleted{$\alpha^3 \xi_6$}} \bm{\mathcal{M}}_0, \\
  &u := \dfrac{1}{2(1+\nu)}.
\end{align}
\end{subequations}
\replaced{where}{Let} $\nu$ denote\added{s} the Poisson's ratio.
\deleted{The potential energy density $\mathcal{U}$ consists of three terms. 
The first term represents the elastic energy associated with extension, shear, and bending along the $s$-axis. 
The second term accounts for the energy resulting from the change in the circumferential length of the infinitesimal element from $r d\theta$ to $\alpha r d\theta$.
The third term corresponds to the energy due to the change in the radial length of the infinitesimal element from $dr$ to $\alpha dr$.}

Whereas a fully coupled 3D elasticity model or a hyperelastic potential (e.g., Neo-Hookean) could \replaced{consider}{rigorously capture} the multi-axial coupling within the strain energy, \added{the stiffness and viscosity matrices in} such approaches \replaced{have}{introduce dense} off-diagonal coupling terms that \replaced{make}{significantly complicate} the analytical derivation of the Euler-Poincaré equations \added{complicated}. 
\replaced{For}{To prioritize} computational efficiency\deleted{ and preserve the mathematically tractable Lie group structure of SIM(3)}, we \deleted{deliberately} employ a simplified macroscopic model \added{preserving the mathematically tractable Lie group structure of SIM(3)}. In \eqref{eq:mathcal{U}}, the strain energy is formulated using a decoupled linear constitutive relation based on SIM(3) kinematics. 
\deleted{The thickness contraction inherent to incompressible elastomers is not handled within the stress-strain tensor itself, but is instead imposed geometrically as an exact volume-preserving constraint \eqref{eq:varepsilon} during the energy integration.}

\subsection{Detailed Formulation of Poincar\'e Equation}
\replaced{We employ}{To account for viscous forces, we adopt} the Kelvin--Voigt model \cite{linn2013geometrically} \added{to represent viscous forces}. 
\deleted{Let $\mu$ denote the viscosity coefficient.}
The \added{viscosity} matrix $\bm{\Upsilon}$ \replaced{can be calculated}{is defined} as follows:
\begin{align}
  \bm{\Upsilon}
  =
  \mu~\mathrm{diag}\{ 3,3,1,1,1,3,1 \} \mbox{\deleted{$\alpha^3 \xi_6$}} \bm{\mathcal{M}}_0
\end{align}
\replaced{where}{Let} $\mu$ denote\added{s} the viscosity coefficient\deleted{ in the Kelvin--Voigt model \cite{linn2013geometrically} for the viscoelastic force}. 

\deleted{To simplify the derivation of the fundamental equations of motion, we initially assume that the mass, stiffness, and viscosity matrices remain constant during deformation. Specifically, we set}
\deleted{where the subscript $0$ denotes their respective constant values in the reference configuration.} 
\deleted{This assumption yields the most mathematically tractable and basic form of the PC7 dynamics. 
A more physically rigorous model tailored for hollow structures such as thin-walled cylinders, which strictly enforces the perfect incompressibility condition ($dV = dV_0$) to precisely account for thickness variations, will be detailed later in Section \ref{sectionW}.}

\replaced{Substituting \eqref{eq:mathcal{K}} and \eqref{eq:mathcal{U}} into}{Under these assumptions, the Poincar\'e equations} \eqref{eq:poincare} \replaced{yields}{are expanded into the following explicit form}
\begin{align}
  \mbox{\replaced{$\widehat{\bm{\mathcal{M}}}$}{$\alpha^2 \bm{\mathcal{M}}_0$}} \dot{\bm{\mathcal{V}}} + \bm{\mathcal{B}}
  &= \bm{\mathcal{F}}
  \label{eq_m_2_sim3}
\end{align}
\begin{subequations}
\begin{align}
  \bm{\mathcal{B}} 
  &:=
  \mbox{\added{$(2 \bm{e}_7^T \bm{\mathcal{V}} \bm{E} - [\bm{\mathcal{V}} \times_7]^T - \bm{e}_7 \bm{\mathcal{V}}^T) \widehat{\bm{\mathcal{M}}}\bm{\mathcal{V}} $}}
  \\
  &\mbox{\deleted{$2 \alpha^2 \mathcal{V}_{\lambda} \bm{\mathcal{M}}_0 \bm{\mathcal{V}}
  -[\bm{\mathcal{V}}\times_7]^T \alpha^2 \bm{\mathcal{M}}_0 \bm{\mathcal{V}}
  - \alpha \bm{\mathcal{V}}^T \alpha\bm{\mathcal{M}}_0 \bm{\mathcal{V}}\bm{e}_7 $}}
  \nonumber
  \\
  \bm{\mathcal{F}} &= \bm{\mathcal{F}}'_{\mathrm{int},\xi}
  - [\bm{\xi}\times_7]^T \bm{\mathcal{F}}_{\mathrm{int},\xi}
  +\bm{\mathcal{F}}_{\mathrm{int},\alpha}
  +\bm{\mathcal{F}}_{\mathrm{ext}}
  \\
  \bm{\mathcal{F}}_{\mathrm{int},\xi} 
  &:=
  \mbox{\replaced{$- \widehat{\bm{\Sigma}}(\bm{\xi} - \alpha^{-1}\bm{\xi}_0) - \widehat{\bm{\Upsilon}}(\bm{e}_7^T\bm{\mathcal{V}} \bm{\xi} + \dot{\bm{\xi}})$}{$-\alpha \bm{\Sigma}_0 (\alpha\bm{\xi}-\bm{\xi}_0)
  -\alpha \bm{\Upsilon}_0 \dot{(\alpha\bm{\xi})}$}}
  \\
  \bm{\mathcal{F}}_{\mathrm{int},\alpha}
  &:=
  \left(\mbox{\replaced{$\bm{\xi}^T \bm{\mathcal{F}}_{{\rm int}, \xi}$}{$- \alpha\bm{\xi}^T \bm{\Sigma}_0 (\alpha\bm{\xi}-\bm{\xi}_0)$}}
  - 2 \alpha E_0 A_0 (\alpha - 1)\right.\notag\\
  &\quad \left. \mbox{\deleted{$- \alpha \bm{\xi}^T \bm{\Upsilon}_0 \dot{(\alpha\bm{\xi})}$}}
  - 2 \alpha \mu_0 A_0 \dot{\alpha} \right)\bm{e}_7
\end{align}
\end{subequations}
\added{where $\bm{\mathcal{F}}_{\mathrm{ext}}$ is a term summarizing external forces including the gravitational force and actuation force (e.g., pressure).}

\begin{figure}[tbp]
  \begin{center}
    \includegraphics[width=0.9\hsize]{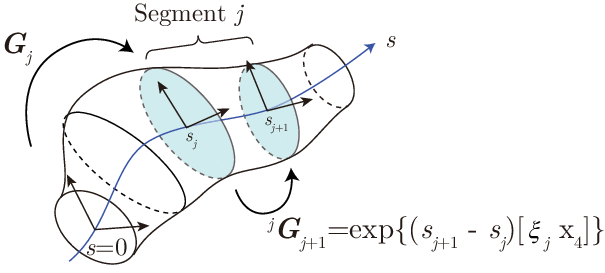}
  \end{center}
  \caption{Piece-wise constant 7D deformation (PC7) model.}
  \label{fig:pc7_model}
\end{figure}
\section{Piece-wise Constant 7D Deformation Model \deleted{(PC7 Model)}\label{sect:pc7_model}}
\subsection{\added{Kinematics of PC7 Model}}
\replaced{In a manner similar to the PCS model \cite{renda2018discrete}}{Following the Piecewise Constant Strain (PCS) approach}, we discretize \replaced{a}{the} microstretch rod into \replaced{finite number of}{$n$} segments\replaced{ and assume that}{, assuming} the \added{7D} deformation vector $\bm{\xi}(t,s)$ \replaced{is}{remains} constant within each segment \deleted{$s \in [s_{i-1}, s_i]$\cite{renda2018discrete}}.
\added{In this paper, we refer to this discrete microstretch rod model as piece-wise constant 7D deformation (PC7) model.}

\added{As shown in \figref{fig:pc7_model}, we define segment $j$ in $s \in [s_{j}, s_{j+1}]$ and the constant deformation vector as}
\begin{align}
    \bm{\xi}_j(t) := \bm{\xi}(t,s) \ (s \in [s_{j}, s_{j+1}]) .
\end{align}
\added{Given $\bm{G}(s_j)$, $\bm{G}(s)$ at any arbitrary $s \in [s_{j}, s_{j+1}]$ can be calculated using the exponential map as}
\begin{align}
    \label{eq:pc7_fk}
    \bm{G}(s) = \bm{G}(s_j) \exp\{ (s - s_j ) [\bm{\xi}_j \times_4]\} \ (s \in [s_{j}, s_{j+1}]) .
\end{align}
\added{This is a direct extension just replacing $\bm{G} \in$ SE(3) with $\bm{G} \in$ SIM(3) in the forward kinematics of the PCS model \cite{renda2018discrete}. 
The common mathematical properties among Lie groups enables this extension.
In $\mathfrak{sim}(3)$, given a 7D vector $\bm{x} = [\bm{x}_a^T \ \bm{x}_{\ell}^T \ \lambda ]^T$, the exponential map can be calculated as}
\begin{align}
    \label{eq:exp_mapping_sim3}
 \exp([\bm{x} \times_4]) = 
 \begin{bmatrix}
  e^{\lambda} \exp([\bm{x}_a \times_3]) & \bm{V} \bm{x}_2\\
  \bm{0}^T & 1
 \end{bmatrix}.
\end{align}
\added{The matrix $\bm{V}$ can be calculated as described in Appendix \ref{apdx:exp}.}

\added{Using \eqref{eq:pc7_fk} with $s = s_{j+1}$, we can recursively calculate $\bm{G}(s_{j+1})$ at the next segment and subsequently every coordinate of the rod.}
\replaced{Therefore, we can define the generalized coordinate concatenating all $\bm{\xi}_j$ as}{The generalized coordinate vector $\bm{q} \in \mathbb{R}^{7n}$ is formed by concatenating these constant 7D strains\cite{renda2018discrete}. }
\begin{align}
    \bm{q}
    =
    \begin{bmatrix}
        \bm{\xi}_1^T & \cdots & \bm{\xi}_N^T
    \end{bmatrix}^T
    \in \mathbb{R}^{7N}
\end{align}
\added{where $N$ is the number of division of segments.}
\replaced{Eventually, we can formulate forward kinematics in a similar manner to traditional rigid link system and also PCS model. In particular, the differential forward kinematics can be represeted as}{Because the SIM(3) formulation preserves the exact Lie group structure of SE(3), the forward and differential kinematics ($\bm{\mathcal{V}}(t,s) = \bm{J}(\bm{q})\dot{\bm{q}}(t)$) can be computed using the identical recursive exponential maps and geometric Jacobian $\bm{J}(\bm{q})$ as the standard PCS model, simply by replacing the $\mathfrak{se}(3)$ operators with their $\mathfrak{sim}(3)$ counterparts\cite{renda2018discrete}.}
\begin{align}
    \label{eq:pc7_diff_fk}
    \bm{\mathcal{V}}(t,s) = \bm{J}(\bm{q})\dot{\bm{q}}(t)
\end{align}
where $\bm{J}(\bm{q})$ is the Jacobian matrix of the PC7 model.

\subsection{\added{Dynamics of PC7 Model}}
We can derive the equation of motion projecting the continuous dynamics \added{\eqref{eq_m_2_sim3}} onto the \deleted{joint} space\added{ of $\bm{q}$} via the extended Hamilton's principle\added{, in a manner similar to} \cite{renda2018discrete}.
\added{The time-derivative of \eqref{eq:pc7_diff_fk} is calculated as}
\begin{align}
    \dot{\bm{\mathcal{V}}} = \bm{J}\ddot{\bm{q}} + \dot{\bm{J}}\dot{\bm{q}}.
\end{align}
\replaced{Substituting this equation into \eqref{eq_m_2_sim3} and integrating it though $s$ yield}{By substituting the differential kinematics into our SIM(3) Poincar\'e equations \eqref{eq_m_2_sim3} ($\alpha^2 \bm{\mathcal{M}}_0 \dot{\bm{\mathcal{V}}} + \bm{\mathcal{B}}=\bm{\mathcal{F}}$), we obtain the generalized equations of motion:}
\begin{align}
    \label{eq:pc7_dynamics}
    \bm{M}(\bm{q})\ddot{\bm{q}} + \bm{b}(\bm{q}, \dot{\bm{q}}) = \bm{\tau},
\end{align}
where \deleted{the generalized mass matrix $\bm{M}$, the bias force vector $\bm{b}$, and the generalized force vector $\bm{\tau}$ are defined as}
\begin{subequations}
\begin{align}
  \bm{M}(\bm{q}) &:= \int_0^{s_n} \bm{J}^T \mbox{\replaced{$\widehat{\bm{\mathcal{M}}}$}{$\alpha^2{\bm{\mathcal{M}}_0}$}} \bm{J} ds, \\
  \bm{b}(\bm{q}, \dot{\bm{q}}) &:= \int_0^{s_n} \bm{J}^T \left( \mbox{\replaced{$\widehat{\bm{\mathcal{M}}}$}{$\alpha^2{\bm{\mathcal{M}}_0}$}} \dot{\bm{J}} \dot{\bm{q}} + \bm{\mathcal{B}} \right)ds, \\
  \bm{\tau} &= \int_0^{s_n} \bm{J}^T \bm{\mathcal{F}} ds.
\end{align}
\end{subequations}

\subsection{\added{Dynamics with Geometric Constraint}}
\added{Similarity of the proposed PC7 model with the rigid-link and PCS models allows us to utilize existing techniques for calculating and simulating various systems.}
\added{One example is simulation under holonomic constraint, which is generally represented by}
\begin{align}
    \label{eq:constraint}
    \bm{f}(\bm{q}) = \mathrm{const} .
\end{align}
\added{This formulation can represent various geometric constraints, e.g., constant volume constraint or shape deformation constraint of McKibben actuator. 
Examples of detailed formulation of the function $\bm{f}(\bm{q})$ will be shown in the next section.}
Under the constraint represented by \eqref{eq:constraint}, the equation of motion including constraint force $\bm{\kappa}$ can be represented as follows \cite{featherstone2014rigid, armanini2021discrete}:
\begin{align}
    \begin{bmatrix}
    \bm{M} & \bm{L}\\
    \bm{L}^T & \bm{0}
  \end{bmatrix}
  \begin{bmatrix}
    \ddot{\bm{q}}\\
    \bm{\kappa}
  \end{bmatrix}
  &=
  \begin{bmatrix}
    \bm{\tau} - \bm{b}\\
    -\dot{\bm{L}}^T \dot{\bm{q}}
  \end{bmatrix}\label{eqcc}, \quad
  \bm{L} := \pdif{\bm{f}}{\bm{q}}.
\end{align}
We can obtain the acceleration $\ddot{\bm{q}}$ solving this equation, and simulate the time-evolution of the deformation integrating $\ddot{\bm{q}}$.

\added{This is one of the advantages of the PC7 model, which allows us to systematically integrate various physical characteristics, especially geometric and kinematic constraints.
In the next section, we formulate such constraints in detail.}

\begin{figure}[tbp]
  \begin{center}
    \includegraphics[width=0.3\hsize]{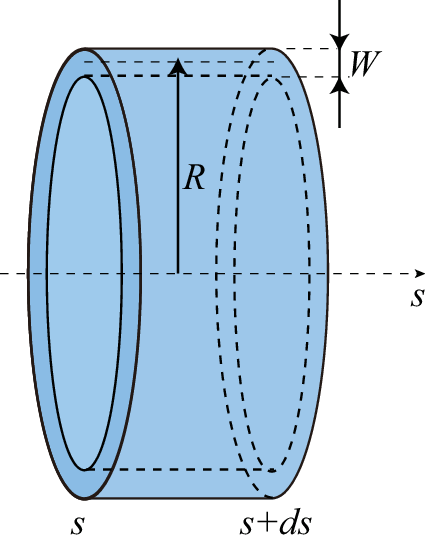}   
  \end{center}
  \caption{Cross section of a thin-walled cylinder.}
  \label{cylinder}
\end{figure}
\section{\replaced{Examples of Constrained Expansion Mechanisms}{Modeling of Constrained Expansion Mechanisms: Thin-Walled Cylinders and McKibben Actuators}\label{sect:ex_constrained_mechanisms}}
\deleted{（前節にまとめて記述するため削除）A key advantage of the proposed PC7 model is its versatility and extensibility. 
The framework allows for the systematic integration of various physical characteristics, specifically: custom assumptions regarding deformation, geometric and kinematic constraints, and actuator-specific force transformations. 
In this section, we demonstrate how this framework accommodates specific physical phenomena---such as kinematic constraints on the wall thickness, the geometric coupling in McKibben artificial muscles, and the mapping of internal fluidic pressure into generalized forces.
To simulate the forward dynamics under specific geometric constraints, we employ the standard constrained dynamics formulation using Lagrange multipliers. For a given set of holonomic constraints $\bm{f}(\bm{q}) = \mathrm{const}$, the constrained equations of motion and the corresponding constraint forces are computed following the established methodologies detailed in \cite{featherstone2014rigid, armanini2021discrete}. The specific definitions of the constraint functions $\bm{f}(\bm{q})$ for each type are formulated in the following subsections.}

\subsection{\added{Basic Geometry}}
\deleted{Before detailing the actuator-specific models, we establish the fundamental geometric properties and boundary conditions shared by both the thin-walled cylinders and the McKibben actuators.}

\added{In this section, we formulate geometric constraints for two examples of expansion mechanisms: thin-walled cylinder and McKibben-type artificial muscle actuator.}
\deleted{Since both mechanisms are fundamentally based on a tubular structure,}  
\figref{cylinder} illustrates an infinitesimal segment of \deleted{such} a cylinder within \deleted{the material interval} $[s, s+ds]$.
\replaced{Let $R(t,s)$ and $W(t,s)$ denote the radius and wall thickness in the deformed configuration at time $t$, respectively.}{In the deformed configuration at time $t$, the cross-section is characterized by its current radius $R(t,s)$ and wall thickness $W(t,s)$. }
\replaced{In particular,}{Conversely, let} $R_0$ and $W_0$ denote the corresponding \replaced{values}{geometric parameters} in the reference configuration. 
\replaced{T}{Based on these initial dimensions, t}he cross-sectional area $A_0$ and the area moments of inertia $J_{x0}$, $J_{y0}$, and $J_{z0}$ in the reference configuration are calculated as follows:
\begin{align}
  A_0 = 2 \pi R_0 W_0, \
  J_{x0} = J_{y0} = \frac{1}{2}R_0^2 A_0, \
  J_{z0} = R_0^2 A_0.
\end{align}

\replaced{I}{Furthermore, i}n practical fabrications, the distal end of a soft \deleted{fluidic} actuator is typically sealed or rigidly clamped by a rigid end-cap\added{ at the tip ($s=s_n$)}. This \deleted{physical restriction prevents any radial expansion or contraction at the tip ($s=s_n$), forcing the terminal cross-section to remain in its reference configuration (i.e., $\alpha = 1$). This boundary} constraint \replaced{can be represented}{is mathematically expressed} as
\begin{align}
    \label{eq:boundary_constraint}
    f_1(\bm{q}) := \alpha(s_n) = 1.
\end{align}
Note that \replaced{in general no explicit constraint is necessary at the base ($s=0$) because such a soft actuator is fixed to a rigid base frame, naturally satisfying $\alpha(0)=1$}{the proximal end ($s=0$) is inherently fixed to the rigid base frame, naturally satisfying $\alpha(0)=1$. Therefore, no explicit constraint equation is required for the base in the generalized coordinate formulation}.

\subsection{Thin-Walled Cylinders\label{sectionW}}
\replaced{In the case of a thin-walled cylinder, the volume change is different from \eqref{eq:volume_change} because of the Poisson's effect. In this subsection, we reformulate it for a thin-walled cylinder including McKibben actuator}{In the microstretch formulation, hollow flexible structures---such as thin-walled cylinders and McKibben actuators---experience significant variations in wall thickness during radial expansion and axial contraction. Therefore, a specific kinematic treatment for the wall thickness is required}.
\replaced{The radial and axial deformations are represented by $\alpha$ and $\alpha \xi_6$, respectively.}{The radial deformation is represented by the isotropic scaling factor $\alpha$, while the axial stretch is represented by $\alpha \xi_6$. }
We assume that an infinitesimal thickness element $dr$ in the reference configuration deforms to $\varepsilon dr$ in the deformed configuration\added{ where $\varepsilon$ is the thickness stretch ratio}. 
\replaced{Using}{By incorporating the Poisson's effect characterized by} the Poisson's ratio $\nu$, \deleted{the wall thickness contracts in response to the radial strain $(\alpha - 1)$ and the axial strain $(\alpha \xi_6 - 1)$. The thickness stretch ratio} $\varepsilon$ \replaced{can be calculated considering the radial strain $(\alpha - 1)$ and axial strain $(\alpha \xi_6 - 1)$}{is mathematically defined} as follows:
\begin{align}
  \varepsilon &= 1 - \nu\left\{ (\alpha - 1) + (\alpha\xi_6 - 1) \right\} \nonumber \\
  &= 1 + 2\nu - \nu\alpha(1 + \xi_6).
\end{align}
\replaced{Different from \eqref{eq:volume_change}, the}{Accordingly, an} infinitesimal volume \replaced{in}{element per unit length of the material space $s$, which is given by $dV_0 = r d\theta dr$ in the reference configuration, is mapped to} the deformed configuration \added{is calculated} as
\begin{subequations}
\begin{align}
  dV = (\alpha r d\theta)(\varepsilon dr)(\alpha \xi_6 \mbox{\added{$ds$}}) = \overline{v} dV_0,
    \\
    \overline{v} := \varepsilon \alpha^2 \xi_6.
\end{align}
\end{subequations}
However, the formulations after \eqref{eq:kinetic_energy} are considered in the reference configuration and do not include the volume change ratio $v$, which allows us to still use the same formulations even when considering a thin-walled cylinder.
\deleted{To account for this thickness variation in the dynamic model, the volume element per unit length in both the kinetic and potential energy densities (\eqref{eq:mathcal{K}} and \eqref{eq:mathcal{U}}) is updated from $\alpha^3 \xi_6 dV_0$ to $\varepsilon \alpha^2 \xi_6 dV_0$. 
Furthermore, the potential energy term corresponding to the change in the radial length of the infinitesimal element is replaced by $\frac{1}{2} E \varepsilon \alpha^2 \xi_6 A_0 (\varepsilon - 1)^2$. 
By substituting this updated Lagrangian density into the SIM(3) Poincar\'e equations, the corresponding internal viscoelastic forces $\bm{\mathcal{F}}_{\mathrm{int},\xi}$ and $\bm{\mathcal{F}}_{\mathrm{int},\alpha}$ are systematically derived. Note that the expanded algebraic expressions for these internal forces are omitted here for brevity, as they can be straightforwardly obtained through standard partial differentiation of the updated Lagrangian.}

\subsection{McKibben Actuators}
\replaced{A McKibben-type artificial
muscle actuator \cite{tondu2002modeling} consists of a thin-walled cylindrical rubber tube and
an outer sleeve in which fibers are braided in a pantograph like structure.
When pneumatic pressure is applied inside the rubber tube, the tube tends to expand. However, the sleeve deforms like a pantograph, allowing the muscle to expand in the radial direction while contracting in the axial direction as the braid angle of the sleeve changes. 
Thus, the radial expansion force generated by the rubber tube is converted into an axial contraction force by the sleeve.}{A McKibben pneumatic artificial muscle comprises\cite{tondu2002modeling} an inner elastomeric tube constrained by an outer sleeve of nearly inextensible braided fibers. 
When internal pneumatic pressure is applied, the inner tube expands radially. 
Because the braided fibers are inextensible, this radial expansion geometrically forces the braid angle to increase, thereby mechanically coupling the radial expansion with an axial contraction. 
Through this pantograph-like kinematic coupling, the radial expansion force generated by the internal pressure is effectively converted into an axial contraction force along the actuator.}

\begin{figure}[tbp]
  \begin{center}
    \includegraphics[width=80mm]{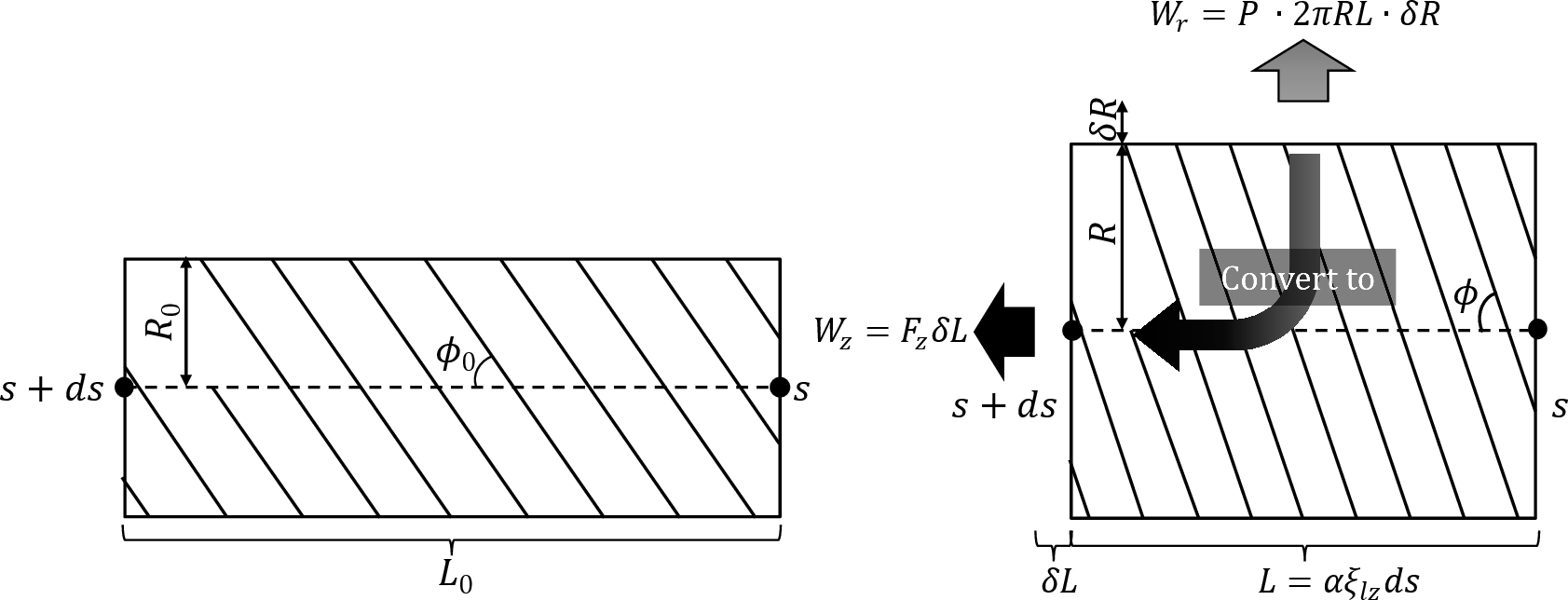}   
  \end{center}
  \caption{Parameters of a McKibben artificial muscle actuator. 
  }
  \label{MK}
\end{figure}
As shown in \figref{MK}, for an infinitesimal segment $[s, s+ds]$, let $\phi(t,s)$ denote the braid angle of the inextensible sleeve, $R(t,s)$ the radius of the artificial muscle, and $L(t,s)$ its length defined per unit material length $s$ in the deformed configuration. Let $\phi_0$, $R_0$, and $L_0=1$ define their corresponding initial values in the reference configuration.
Due to the geometric constraints of the braided sleeve \cite{tondu2012modelling}, the kinematic deformation satisfies the following relationships:
\begin{align}
  \frac{R}{R_0} = \alpha = \frac{\sin\phi}{\sin\phi_0}, \quad
  \frac{L}{L_0} = \alpha\xi_6 = \frac{\cos\phi}{\cos\phi_0}.
\end{align}
In addition to the boundary constraint \replaced{given by \eqref{eq:boundary_constraint}}{$f_1(\bm{q}) = \alpha(s_n) = 1$ defined previously}, \deleted{this strict kinematic coupling between the radial and axial deformations imposes} the following invariant constraint \added{is imposed} on the generalized coordinates:
\begin{align}
    \label{eq:mckibben_constraint}
  f_2(\bm{q}) = \alpha^2 \sin^2\phi_0 + \alpha^2 \xi_6^2 \cos^2\phi_0 = 1.
\end{align}
Furthermore, to maintain consistency in the radial deformation under the piece-wise constant assumption, we constrain the spatial derivative of the scaling factor, $\bm{\xi}_i^T \bm{e}_7$, for the intermediate segments ($i \neq 1, n$). This constraint is represented as:
\begin{align}
  f_{3,i}(\bm{q}) = \bm{\xi}_i^T \bm{e}_7 = 0.
\end{align}
Physically, this \added{constraint} ensures that the radial expansion $\alpha$ remains uniform within each intermediate segment. If $\alpha$ were to vary spatially ($\bm{\xi}_i^T \bm{e}_7 \neq 0$), the \replaced{constraint given by \eqref{eq:mckibben_constraint}}{invariant condition $f_2(\bm{q}) = 1$} could not hold consistently across the segment, given that the axial strain $\xi_6$ is assumed to be constant. 
\replaced{Note that}{Conversely,} this \deleted{uniform radius} constraint is not applied to the first ($i=1$) and last ($i=n$) segments\replaced{ because these segments}{. Because their outermost boundaries} are rigidly fixed to the reference configuration ($\alpha=1$), \replaced{represented as \eqref{eq:boundary_constraint}.}{by $f_1(\bm{q})$, these terminal segments must accommodate a spatial transition, resulting in a gradual radial expansion from the base and a corresponding contraction toward the tip.}

\section{Simulation\label{sect:simulation}}
\subsection{Simulation of Thin-Walled Cylinder}
We \replaced{verify the validity of}{performed a forward dynamics simulation of a thin cylinder using} the PC7 model \added{via forward dynamics simulations}.
\added{In the following simulations, we prototyped the forward dynamics of the PC7 model using MATLAB and visualized the result of the simulations by another program written using C++.}

\begin{figure}[t]
  \begin{center}
    \begin{subfigure}[b]{0.48\textwidth}
    \includegraphics[width=1.0\hsize]{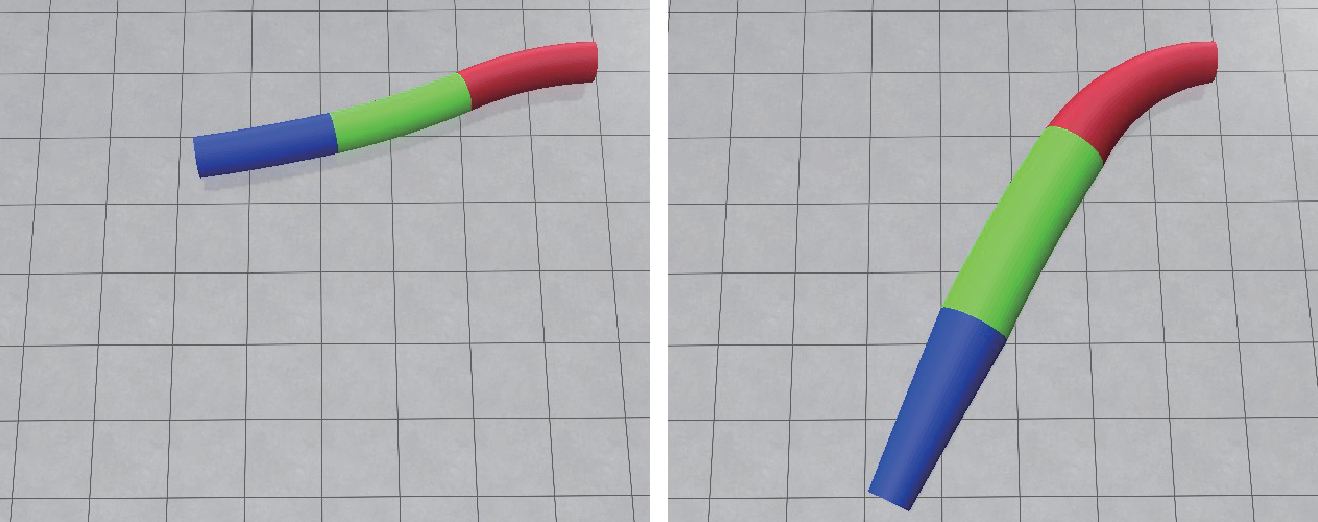}   
    \caption{Snapshots of the simulation\label{gx_sim}}
    \end{subfigure}
    \begin{subfigure}[b]{0.48\textwidth}
    \includegraphics[width=1.0\hsize]{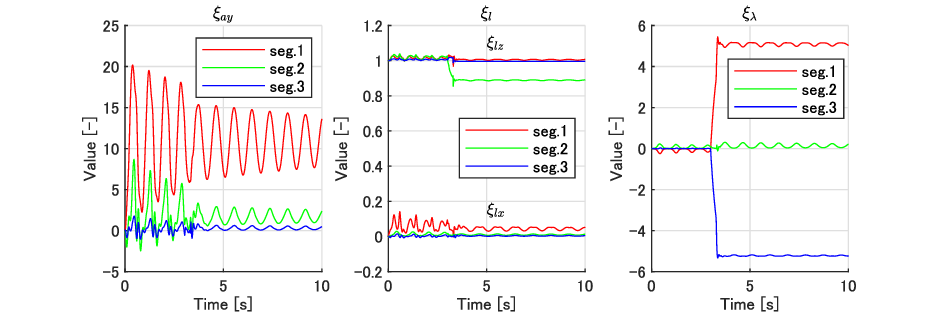}
    \caption{Time-variations of the deformation\label{gx_strain}}
    \end{subfigure}
    \end{center}
  \caption{Results of dynamical simulation of the PC7 model assuming a thin cylinder made of silicone rubber.\label{fig:thin_cylinder_260710_gx}}
\end{figure}

\added{In this section, we firstly performed a simulation assuming a thin-walled cylinder.
Table \ref{param} shows the parameter setting in the simulation. 
The parameters of the material properties were set assuming silicone rubber.}
Figure~\ref{fig:thin_cylinder_260710_gx} shows the simulation result.
As shown in the snapshots (Fig. \ref{gx_sim}), the cylinder was fixed at the base ($s=0$) like a cantilever beam and deformed downward by the gravity effect.
We set a step input of the internal pressure 0 kPa at $t<3$ and 6 kPa at $t \ge 3$, and observed that the cylinder expanded by the pressure.
As shown in Fig. \ref{gx_strain}, the segments 1 and 3 have the same scaling value because of the given constraint.

Note that the thickness of the cylinder assumed in this simulation is 1.5 mm, which is so small that we cannot achieve a stable simulation when using an FEM (Abaqus).
\begin{table}[tbp]
  \begin{center}
  \caption{Parameter setting of thin-walled cylinder assumed in the simulation}
  \label{param}
  \begin{tabular}{c|c}\hline
    Parameter    & Value \\\hline
    Number of segments$~n$ & 3 \\\hline
    $s_1, s_2, s_2$ & $0.1, 0.2, 0.3~$m        \\\hline
    Density$~\rho$ & $1.1\times 10^3~$kg$\cdot$m$^{-3}$ \\\hline
    Initial radius$~R_0$ & $0.015~$m \\\hline
    Initial thickness$~w_0$ & $1.5\times 10^{-3}~$m \\\hline
    Young's modulus$~E$ & $110~$kPa \\\hline
    Poisson's ratio$~\nu$ & $0.5$ \\\hline
    Shear viscosity$~\mu$ & $300~$Pa$\cdot$s \\\hline
  \end{tabular}
  \end{center}
\end{table}

\subsection{Simulation of McKibben Artificial Muscle Actuator}
\begin{figure}[tbp]
  \begin{center}
    \begin{subfigure}[b]{0.48\textwidth}
    \includegraphics[width=1.0\hsize]{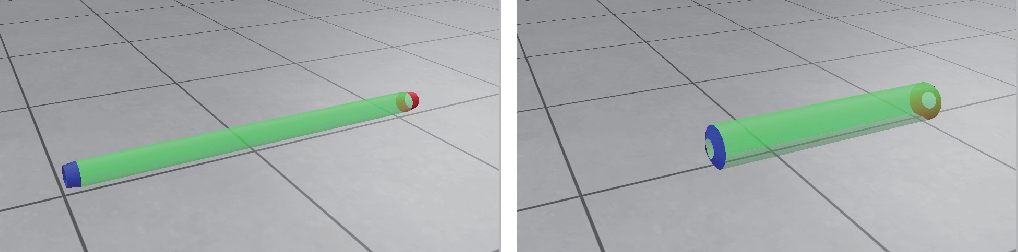}   
  \caption{Snapshots of the simulation\label{fig:mk_sim_snapshots}}
    \end{subfigure}
    \begin{subfigure}[b]{0.48\textwidth}
    \includegraphics[width=\hsize]{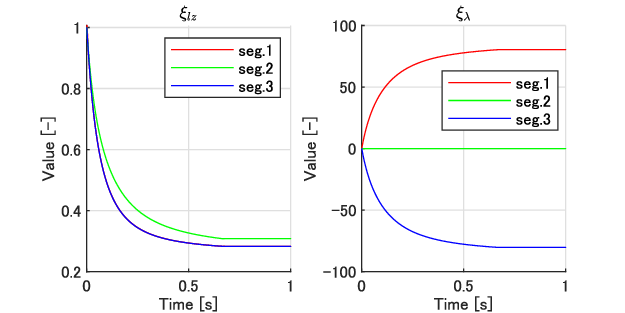}   
  \caption{Time change of the strain vector of the McKibben model.}
  \label{MK_20_3bar_0}
    \end{subfigure}
    \caption{Results of dynamical simulation of the PC7 model assuming a Mckibben-type artificial muscle actuator\label{mk_sim}}
\end{center}
\end{figure}

Next, we conducted a forward dynamics simulation of the McKibben-type artificial muscle actuator.
Table \ref{MKparam} lists the parameter setting used in the simulation.
Figure \ref{mk_sim} shows the simulation result in the case of the braid angle of $\phi_0=20^\circ$ and the pressure of $P=0.3$ MPa.
%
\begin{table}[tbp]
  \begin{center}
  \caption{Parameter Setting of McKibben-type Artificial Muscle Actuator assumed in the Simulation}
  \label{MKparam}
  \begin{tabular}{c|c|c}\hline
    & Parameter    & Value \\ \hline \hline
    & Number of segments$~n$ & 3 \\
    & $s_1, s_2, s_3$ & $0.01, 0.28, 0.01~$m        \\\hline
    \multirow{6}{*}{Rubber} & Density $~\rho$ & $1.1\times 10^3~$kg$\cdot$m$^{-3}$ \\
    & Initial radius$~R_0$ & $7\times 10^{-3}~$m \\
    & Initial thickness$~w_0$ & $0.7\times 10^{-3}~$m \\
    & Young's modulus$~E$ & $1.62~$MPa \\
    & Poisson's ratio$~\nu$ & $0.5$ \\
    & Shear viscosity$~\mu$ & $1~$kPa$\cdot$s \\\hline
    \multirow{6}{*}{Sleeve} & Density$~\rho$ & $1.1\times 10^3~$kg$\cdot$m$^{-3}$ \\
    & Initial radius$~R_0$ & $7\times 10^{-3}~$m \\
    & Initial thickness$~w_0$ & $0.5\times 10^{-3}~$m \\
    & Young's modulus$~E$ & $2~$GPa \\
    & Poisson's ratio$~\nu$ & $0.35$ \\
    & Shear viscosity$~\mu$ & $0~$Pa$\cdot$s \\\hline
    \multirow{6}{*}{Conditions} & Initial weave angle$~\phi_0$ & $20$,$30^\circ$ \\
    & Pressure$~P$ & $0.1$,$0.3$,$0.5$MPa \\
    & Gas constant$~R$ & $287~$J$\cdot$(kg$\cdot$K)$^{-1}$\\
    & Temperature$~T$ & $293~$ K\\
    & Mass flow rate$~\dot{m}$ & $1\times10^{-1}~$kg$\cdot$s$^{-1}$\\\hline
  \end{tabular}
  \end{center}
\end{table}
Figure~\ref{fig:mk_sim_snapshots} shows snapshots of the simulation.
It was observed that the artificial muscle contracted in the axial direction with expanding in the radial direction.
This deformation satisfied the given constraint such that $\xi_7 := \bm{e}_7^T \bm{\xi}$ of the segment $2$ remained zero, and that $\xi_7$ of the segments $1$ and $3$ had the same value.
As in the time response shown in Fig. \ref{MK_20_3bar_0}, it can be seen that the deformation was completed and became stable approximately between $0.4$~s and $0.7$~s.

\begin{figure}[tbp]
  \begin{center}
    \includegraphics[width=0.8\hsize]{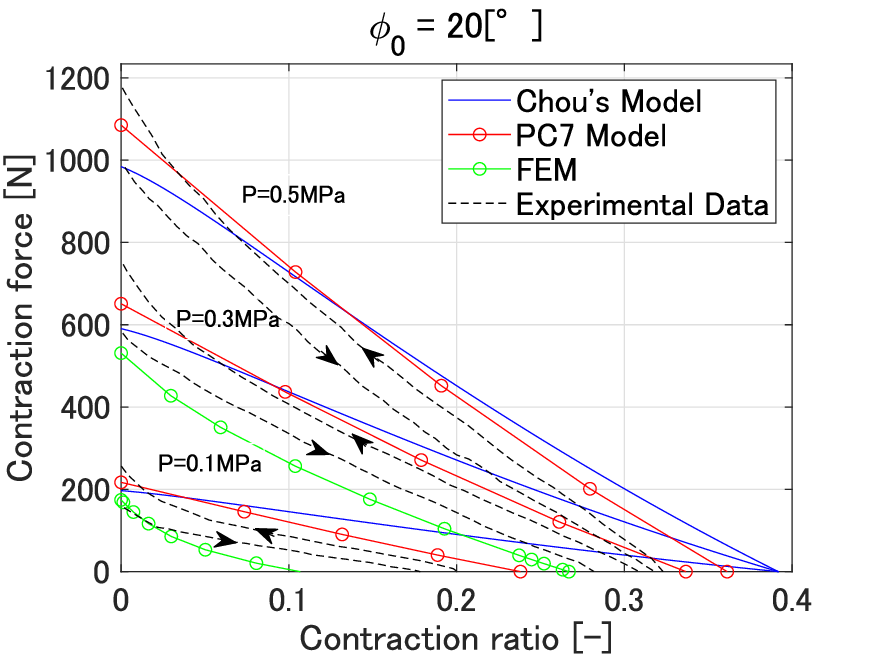}   
  \end{center}
  \caption{Relationships between the contraction ratio and contraction force for an initial weave angle is 20 degrees.}
  \label{forcegraph_20}
\end{figure}
\begin{figure}[tbp]
  \begin{center}
    \includegraphics[width=\hsize]{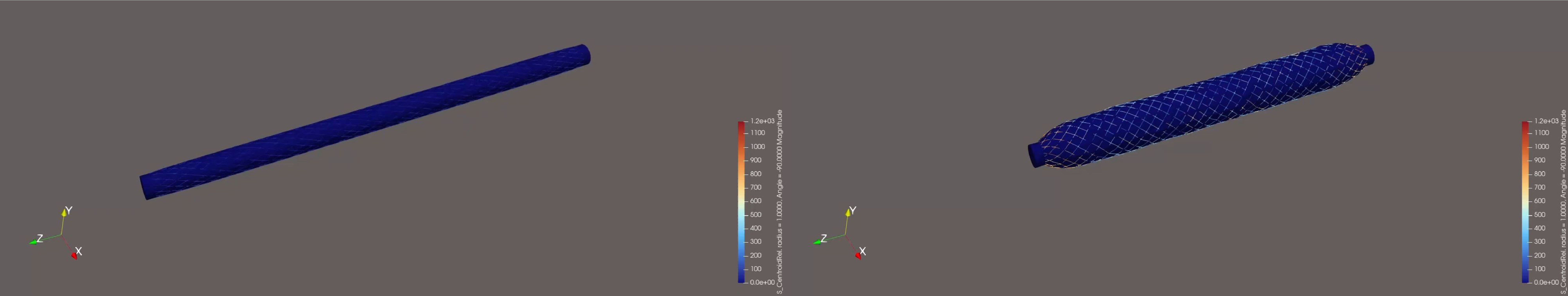}   
  \end{center}
  \caption{Snapshots of the finite element analysis. Before transformation on the left and after transformation on the right.}
  \label{MK_FEM}
\end{figure}

We compared the simulation result by the PC7 model with those based on Chou et al.~\cite{Chou1996} and FEM, and also with the experimental result reported in \cite{tondu2012modelling}.
Figure \ref{forcegraph_20} shows the relationship between the contraction ratio $\varepsilon$ and contraction force $F_c$ obtained in the simulations for three pressure conditions: 0.1, 0.3 and 0.5 MPa.
\begin{itemize}
    \item In the case of the PC7 model (indicated by the red line in Fig. \ref{forcegraph_20}), the contraction force $F_c$ was defined as the axial force applied to the endpoint $s=s_n$ that is required to maintain equilibrium at a given contraction ratio $\Lambda$ calculated as 
\begin{align}
   \Lambda = 1-\frac{\int_{0}^{s_n} L ds}{\int_{0}^{s_n} L_0 ds} = 1-\frac{1}{s_n} \int_{0}^{s_n} \alpha\xi_{6} ds.
\end{align}
    \item In the case of the FEM (indicated by the green line), we used Abaqus and modeled the actuator based on \cite{nozaki2014motion}.
    Figure \ref{MK_FEM} shows snapshots of the FEM simulation.
    \item In Fig.~\ref{forcegraph_20}, we also plotted of the experimental result in \cite{tondu2012modelling} as the ground truth (indicated by the dotted line), extracting the corresponding values using WebPlotDigitizer\footnote{https://automeris.io/}.
\end{itemize}
Note that the FEM computation diverged in the case of $P=0.5$ MPa, which implies it is difficult for the FEM to simulate a rapid deformation due to such a high pressure.

In the simple model by Chou et al. \cite{Chou1996} (indicated by the blue line in Fig. \ref{forcegraph_20}), the maximum contraction ratio at which the contraction force became zero reached the same value regardless of the different pressure conditions.
In contrast, in the PC7 model, the maximum contraction ratio depended on the applied pressure condition as well as FEM.
We consider that this difference stemmed from the consideration of segment discretization and viscoelastic forces in the PC7 model, which were not taken into account in Chou et al. \cite{Chou1996}.

\begin{table}[tbp]
  \centering
  \caption{RMSE results in the condition $\phi_0 =$ 20 deg}
  \label{tab:rmse_20deg}
  \begin{tabular}{llrrr}
    \hline
    Deformation & Pressure & Chou et al. \cite{Chou1996} & PC7 Model & FEM \\
    process &  [MPa] & & & \\
    \hline
    Contruction & 0.5 & 136.0269 & {\bf 120.8651} & N/A \\
                 & 0.3 & 111.1384 & 84.9066  & {\bf 56.9392} \\
                 & 0.1 & 90.0267  & 69.6588  & {\bf 31.9533} \\
    \hline
    Extension & 0.5 & 92.6370 & {\bf 46.5258} & N/A \\
               & 0.3 & 73.3131 & {\bf 31.5580} & 142.4767 \\
               & 0.1 & 60.7822 & {\bf 30.9415} & 77.4292 \\
    \hline
    All & 0.5 & 120.1016 & {\bf 97.4301} & N/A \\
            & 0.3 & 96.2326  & {\bf 66.7565} & 101.7319 \\
            & 0.1 & 78.2038  & {\bf 55.6399} & 56.1502 \\
    \hline
  \end{tabular}
\end{table}
Table \ref{tab:rmse_20deg} summarizes the root mean square errors (RMSEs) of the simulation results using Chou et al. \cite{Chou1996}, PC7 model and FEM, regarding the experimental result reported in \cite{tondu2012modelling}.
It was found that the PC7 model achieved the smalllest RMSE in most cases, excluding the contraction process with $P =$ 0.1 and 0.3 MPa.

Another advantage of the PC7 model is a fast computation compared to the FEM.
The computational time using PC7 model was 26 s whereas the FEM took 109 s (intel Core i9-13900H, 32GB).
These results imply that the PC7 model achieved approximately four times faster computation.
\section{Conclusion}
In this paper, we proposed the PC7 model to consider expanding and contracting deformation perpendicular to the rod central axis.
The obtained results are summarized as follows:
\begin{enumerate}
  \item From mathematical property of SIM(3), we newly defined 7D deformation vector as Lie algebra $\mathfrak{sim}(3)$, which is a direct extension of the 6D deformation vector of the previous PCS model.
  Then, we formulated dynamical equation of the PC7 model based on Poincar\'e equation of microstretch rod.
  \item We simulated an expanding motion of a thin-walled cylinder to validate the PC7 model. 
    We set the thickness of the cylinder to 1.5 mm, which was a severe condition such that the FEM could not perform numerically stable simulation.
    The PC7 model successfully simulated an expanding motion of the thin-walled cylinder.
  \item We also simulated a deformation of McKibben-type artificial muscle actuator using the PC7 model.
  For the quantitative validation, we compare the contraction ratio and force profiles obtained the PC7 model, FEM and simple model with experimental data reported in the previous study.
      The PC7 model achieved the smallest RMSE, and the computational time was 26 s in the case of the PC7 model whereas 109 s in the case of the FEM.
  Also, the PC7 model successfully simulated even with high pressure 0.5 MPa whereas the FEM calculation diverged in the same condition.
\end{enumerate}
It is expected that the PC7 model can be applied not only to artificial muscle actuator but also to volumetric simulation of human muscles.
One of future works is to achieve real-time computation of the PC7 model.
Also, we consider that we can directly extend the analytical derivative framework of the PCS model \cite{ishigaki2024comprehensive} to that of PC7 model, utilizing the similarities found in SE(3) and SIM(3).


\appendices
\section{Calculation of Exponential Map\label{apdx:exp}}
If $\lambda \neq 0$, the matrix $\bm{V}$ in \eqref{eq:exp_mapping_sim3} is calculated as follows:
\begin{subequations}
\begin{align}
 & \bm{V}(\bm{x}_a, \lambda, t) 
 \nonumber
 \\
 &=
 \begin{cases}
  a \bm{E} + b [\bm{x}_a\times] + c[\bm{x}_a\times]^2 & \mbox{if} \ \theta \neq 0, \lambda \neq 0
  \\
  a \bm{E} & \mbox{if} \ \theta = 0, \lambda \neq 0
 \end{cases},
 \\
 a &= \dfrac{e^{\lambda t}-1}{\lambda},
 \\
 b &= \frac{\theta (1-e^{\lambda t} \cos(\theta t)) + e^{\lambda t} \lambda \sin(\theta t)}{\theta (\lambda^2 + \theta^2)},
 \\
 c &=
 \frac{e^{\lambda t} - 1}{\lambda \theta^2} - \frac{e^{\lambda t} \sin (\theta t)}{\theta (\lambda^2 + \theta^2)}
 -
 \frac{\lambda (e^{\lambda t} \cos (\theta t) - 1)}{\theta^2 (\lambda^2 + \theta^2)}.
\end{align}
\end{subequations}




\bibliographystyle{IEEEtran}
\bibliography{IEEEabrv,reference}

\end{document}